%% file: main.tex
\documentclass[10pt,twocolumn,letterpaper]{article}

\usepackage{cvpr}              
\usepackage[accsupp]{axessibility}  
\input{preamble}
\definecolor{cvprblue}{rgb}{0.21,0.49,0.74}
\usepackage[pagebackref,breaklinks,colorlinks,allcolors=cvprblue]{hyperref}
\usepackage{multirow} 
\usepackage{graphicx}
\usepackage[export]{adjustbox}

\usepackage{xparse} 
\usepackage{amsmath}
\usepackage{soul, color, xcolor}
\usepackage{multirow}
\usepackage[table]{xcolor}
\usepackage[most]{tcolorbox}
\usepackage{amssymb} 
\usepackage{pifont}
\usepackage{hyperref}
\usepackage{url}
\usepackage{makecell}
\usepackage{algpseudocode}
\usepackage{threeparttable}
\usepackage{booktabs}
\usepackage{graphicx}
\usepackage{subcaption}
\usepackage{tablefootnote} 
\usepackage{xspace}
\usepackage{lipsum}
\usepackage{colortbl}
\usepackage{enumitem}
\usepackage{array}
 \usepackage[utf8]{inputenc}
\usepackage[table,xcdraw]{xcolor}
\usepackage{tabularx} 
\usepackage{balance}
\definecolor{ForestGreen}{rgb}{0.13,0.55,0.13}
\definecolor{DarkRed}{rgb}{0.8, 0.0, 0.0}

\newcommand{\cmark}{\textcolor{green}{\ding{51}}}
\newcommand{\gaincell}[2]{\makecell[c]{#1 \\ \textcolor{ForestGreen}{$\uparrow$#2}}}
\newcommand{\declinecell}[1]{\textcolor{DarkRed}{$\downarrow$#1}}
\newcommand{\xmark}{\textcolor{red}{\ding{55}}}
\newcolumntype{L}{>{\raggedright\arraybackslash}p{3.0cm}}
\newcolumntype{C}{>{\raggedright\arraybackslash}X}

\newtcolorbox{mybox}[2][]{%
  enhanced,
  width=\linewidth, 
  boxrule=0.5pt,     
  arc=2mm,         
  colback=gray!10,   
  colframe=gray!50,  
  coltitle=white,   
  colbacktitle=gray!70, 
  fonttitle=\bfseries,
  title=#2,
  #1
}
\usepackage{multicol}
\usepackage{graphicx}
\usepackage{tcolorbox}
\usepackage{tabularx}
\usepackage{url}
\def\paperID{}
\def\confName{CVPR}
\def\confYear{2026}

\title{
\begin{minipage}{.09\textwidth}
\vspace{-0.5cm}
\centering
\includegraphics[width=2.2\linewidth]{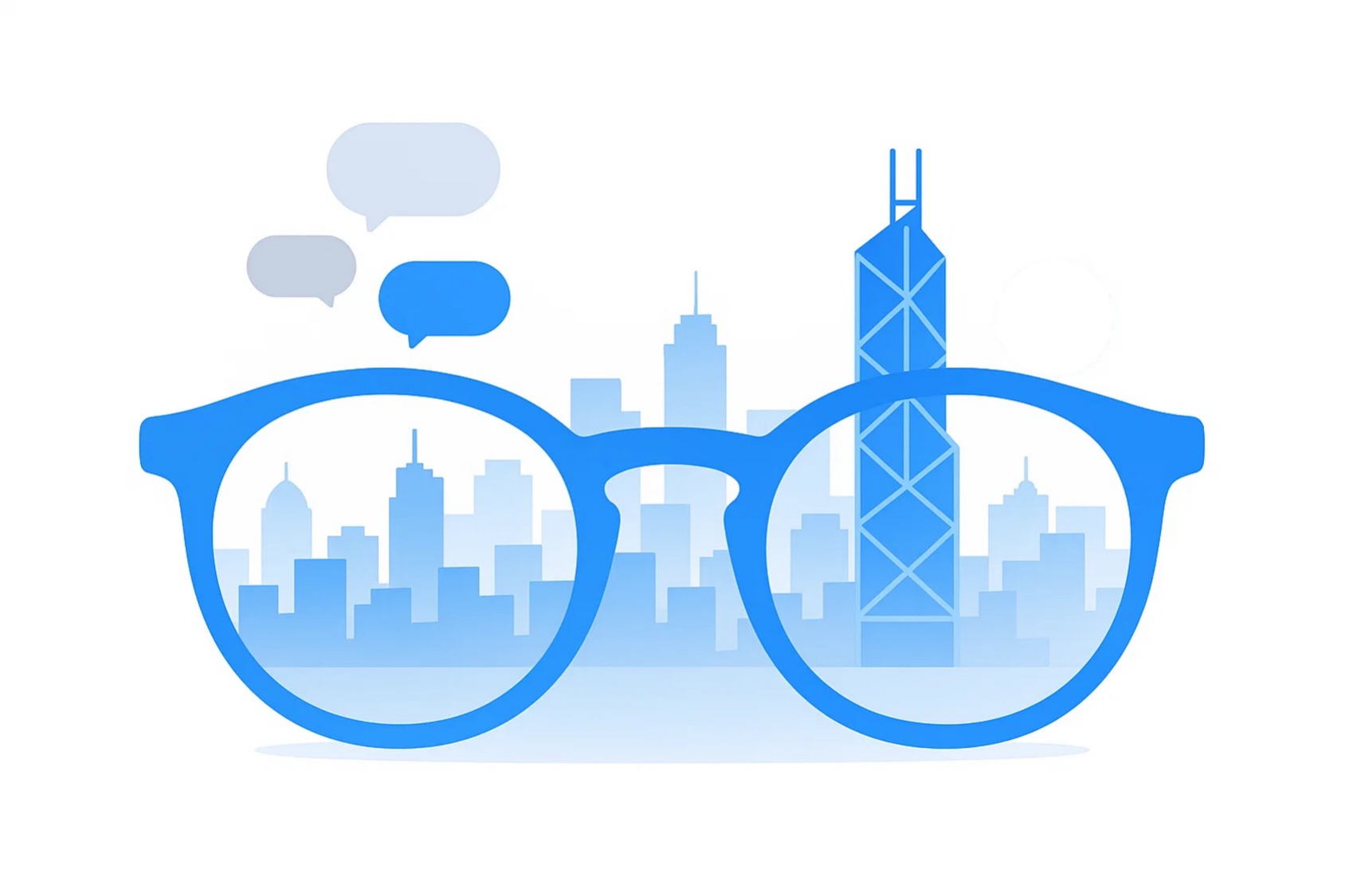}
\end{minipage}%
\hspace{1.2cm}
\begin{minipage}{.8\textwidth}
\centering
\ourname: Benchmarking Vision Language Models as Intelligent Agents for AI Smart Glasses
\end{minipage}}

\author{
Zhuohang Jiang\thanks{Equal contribution.}, \ \ 
Xu Yuan\footnotemark[1], \ \ 
Haohao Qu, \ \ 
Shanru Lin, \ \ 
Kanglong Liu, \\
Wenqi Fan\thanks{Corresponding Authors: Wenqi Fan and Qing Li}, \ \ 
Qing Li\footnotemark[2] \\ [0.1cm]
\small{
   The Hong Kong Polytechnic University 
   } \\
\small{
    \{zhuohang.jiang, xander.yuan, haohao.qu\}@connect.polyu.hk, \{lllam32316, wenqifan03\}@gmail.com
    }\\
\small{
    kl.liu@polyu.edu.hk, csqli@comp.polyu.edu.hk
    }
}

\newcommand{\ourname}{{\textsc{SuperGlasses}}}
\newcommand{\benchname}{{\textsc{SuperGlasses}}}
\newcommand{\agentname}{{\textsc{SuperLens}}}
\newcommand{\collectorname}{{A$^4$}}
\newcommand{\querynum}{{2,422}}

\begin{document}
\maketitle

\input{sections/Abstract}

\input{sections/0_Introduction}

\input{sections/2_GlassesBench}

\input{sections/3_SuperGlass_Agent}
\input{sections/4_Experiment}
\input{sections/5_Conclusion}
\input{sections/6_Acknowledgement}

{
    \small
    
    \bibliographystyle{ieeenat_fullname}
    \bibliography{main}
}

\clearpage
\appendix
\input{sections/9_Appendix}

\end{document}

%% file: sections/Abstract.tex
\begin{abstract}
The rapid advancement of AI-powered smart glasses—one of the hottest wearable devices—has unlocked new frontiers for multimodal interaction, with Visual Question Answering (VQA) over external knowledge sources emerging as a core application. Existing Vision Language Models (VLMs) adapted to smart glasses are typically trained and evaluated on traditional multimodal datasets; however, these datasets lack the variety and realism needed to reflect smart glasses usage scenarios and diverge from their specific challenges, where accurately identifying the object of interest must precede any external knowledge retrieval. 
To bridge this gap, we introduce \textbf{\ourname{}}, the first comprehensive VQA benchmark built on real-world data entirely collected by smart glasses devices. \ourname{} comprises \querynum{} egocentric image-question pairs spanning 14 image domains and 8 query categories, enriched with full search trajectories and reasoning annotations. We evaluate 26 representative VLMs on this benchmark, revealing significant performance gaps. 
To address the limitations of existing models, we further propose the \textbf{\agentname{}}, a multimodal smart glasses agent that enables retrieval-augmented answer generation by integrating automatic object detection, query decoupling, and multimodal web search. 
\textbf{\agentname{}} achieves state-of-the-art performance, outperforming GPT-4o by 2.19\%, underscoring the need for task-specific solutions in smart glasses VQA. Our dataset is publicly available at \href{https://huggingface.co/datasets/xandery/SuperGlasses}{https://huggingface.co/datasets/xandery/SuperGlasses}.

\end{abstract}

%% file: sections/0_Introduction.tex
\begin{figure*}[t]
    \vskip -0.3in
    \centering
    \includegraphics[width=\linewidth]{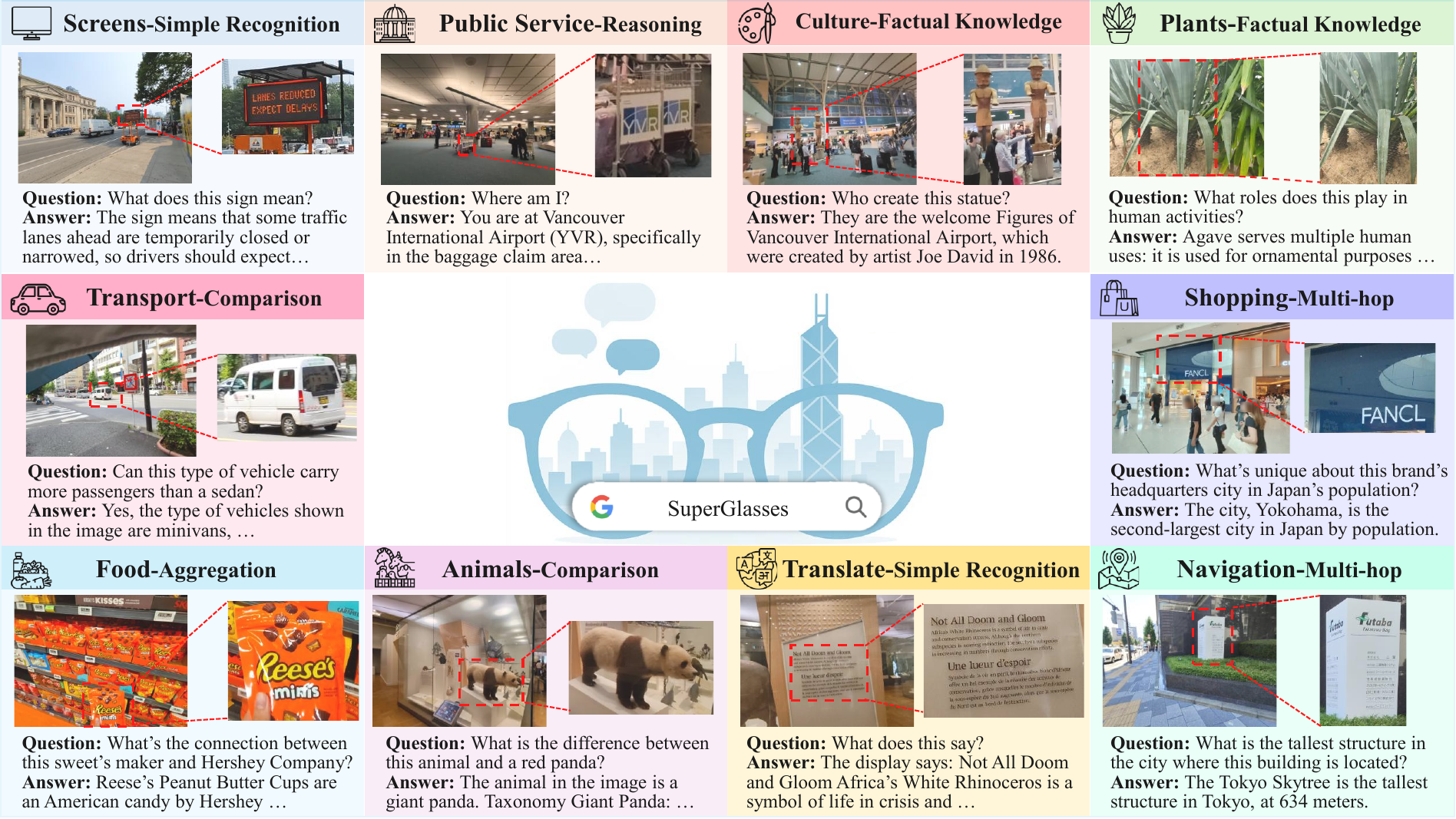}
    \caption{\textbf{\benchname}~contains 14 image domains and 8 query categories, with \querynum{} question–answer pairs. Each example includes an egocentric image captured by smart glasses, a manually annotated question–answer pair, and associated multimodal search logs.}
    \label{fig:dataset_intro}
    \vskip -0.1in
\end{figure*}

\section{Introduction}
Recent advancements in wearable technologies have positioned \textbf{AI-powered Smart Glasses} as a major catalyst for transforming daily human–computer interactions across various domains~\citep{danielsson2020augmented,wang2023practical,chang2025wearvqa,lin2025wearvox}, such as telemedicine support in healthcare~\citep{wang2025systematic} and distance learning and assistance in education~\citep{spitzer2018distance}.
Driven by progress in AI techniques, particularly Large Language Models (\textbf{LLMs})~\citep{brown2020language,ouyang2022training,touvron2023llama,chang2024survey,wang2024survey} and Vision Language Models (\textbf{VLMs})~\citep{yin2024survey,zhu2025internvl3,bai2025qwen25vl,hong2025glm}, modern smart glasses are now capable of seamlessly integrating contextual digital information into the user's visual field, thereby enhancing situational awareness and enables intuitive, natural interactions with the surrounding environment.
The \emph{Ray-Ban Meta AI Glasses}\footnote{\label{fn:rayban}https://www.meta.com/ai-glasses/ray-ban-meta/}, widely regarded as one of the most advanced smart glasses, have surpassed one million units in sales within a few months of their launch, exemplifying the growing consumer adoption and practical relevance of this technology.
As adoption accelerates, such smart wearable devices are poised to redefine the boundaries between digital augmentation and physical reality, marking a pivotal step in the evolution of pervasive, context-aware computing. 
A critical component of smart glasses is the built-in intelligent agent~\cite{cheng2024exploring,xie2024large,ke2024hydra,marsili2025visual,suris2023vipergpt}, which processes information from both users and onboard sensors, effectively functioning as the device's brain. 
Most existing intelligent agents in smart glasses are powered by \emph{large VLMs equipped with a Retrieval-Augmented Generation (RAG) pipeline}~\citep{lewis2020retrieval,rao2023retrieval,fan2024survey}.
Currently, the development and evaluation of these intelligent agents commonly rely on benchmarks from the retrieval-augmented Visual Question Answering (VQA) domain~\cite{wang2023filling,schwenk2022okvqa,lin2022retrieval,wang2025cragmm}, which provide standardized tasks for assessing the models' ability to interpret visual and textual context, retrieve relevant external knowledge, and generate accurate responses to user queries.
For instance, benchmarks such as Dyn-VQA~\cite{liu2025benchmarking}, LIVEVQA~\cite{fu2025livevqa}, CRAG-MM~\cite{wang2025cragmm}, and WearVQA~\cite{chang2025wearvqa} present dynamic questions that are especially challenging due to their rapidly evolving answers, the need for comprehensive multimodal knowledge, and the demand for complex multi-hop reasoning. 

Despite these progresses, a substantial gap remains between the typical VQA tasks focused on by these benchmarks and the real-world demands of AI smart glasses.
\textbf{First}, most images and answers in existing VQA datasets are not sourced from real smart glasses usage scenarios, leading to a significant \emph{scenario gap} when developing and evaluating intelligent agents for smart glasses-oriented systems. 
\textbf{Second}, images in previous VQA benchmarks are typically clear and object-centric, with target items explicitly visible and easy to identify. 
In contrast, as illustrated in Figure~\ref{fig:dataset_intro}, real-world images captured by smart glasses often include large amounts of irrelevant background noise due to their unique visual perception mechanism. 
Consequently, it is crucial for the agents to first detect and localize the target objects before performing question answering, which makes the task substantially more challenging.
\textbf{Finally}, most existing benchmarks lack detailed search records or tool-use trajectories within the RAG paradigm.
However, such traceability is essential for understanding agent behavior in AI-powered smart glasses.
Therefore, there is an urgent need for a comprehensive benchmark that faithfully reflects the practical usage scenarios of smart glasses.

\begin{figure*}[t]
    \vskip -0.3in
    \centering
    \includegraphics[width=\linewidth]{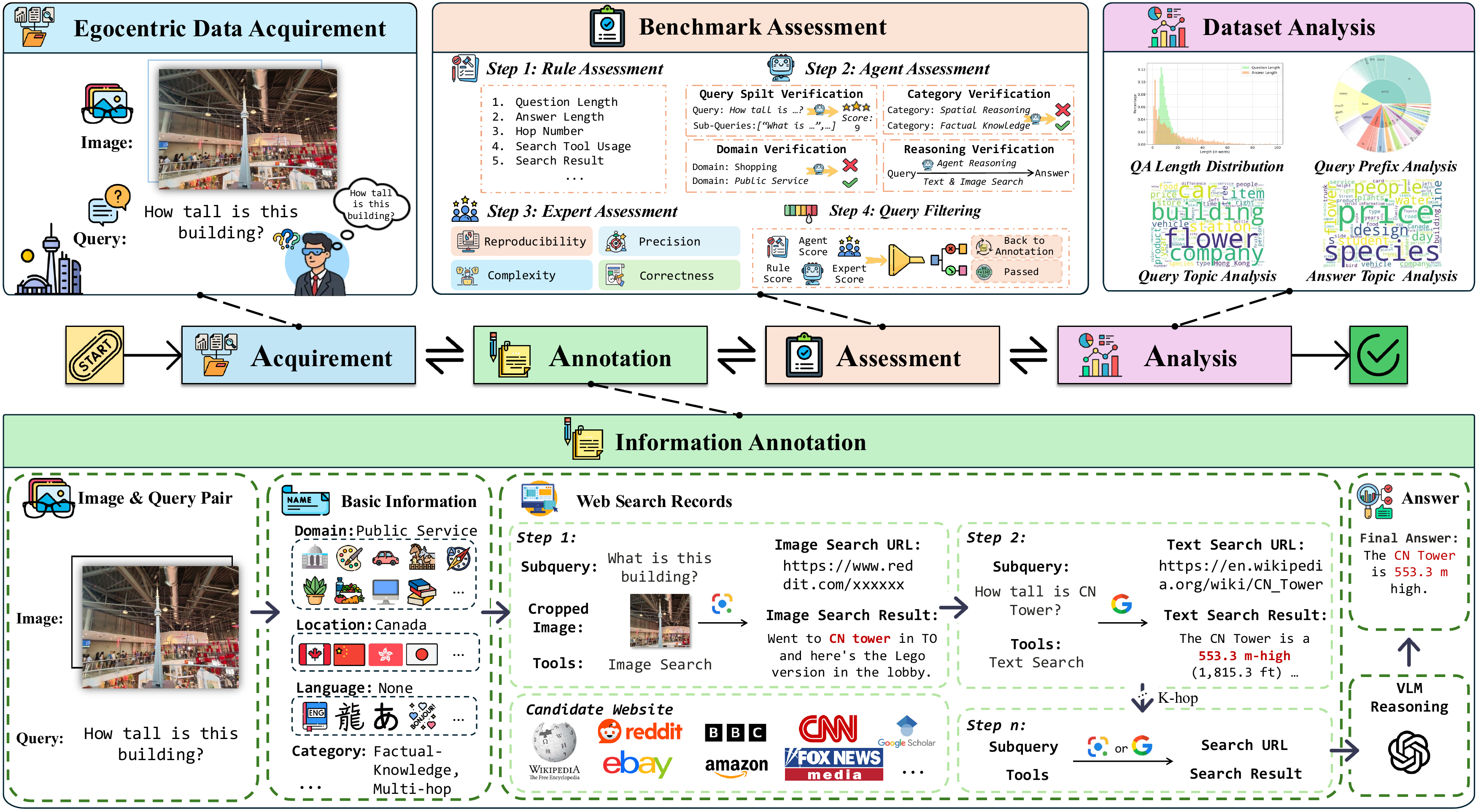}
    \caption{The \textbf{\collectorname{} Data Collection Pipeline}, consisting of four stages: \textbf{Acquirement}, \textbf{Annotatation}, \textbf{Assessment}, and \textbf{Analysis}. }
    \label{fig:a4_pipeline}
    \vskip -0.1in
\end{figure*} 

To bridge these gaps, we propose \textbf{\ourname{}}, a retrieval-augmented VQA dataset tailored for benchmarking VLM agents across diverse real-world smart glasses scenarios.
Our benchmark is the first of its kind, comprising \querynum{} images manually collected using three representative smart glasses (i.e., Ray-Ban Meta$^{\ref{fn:rayban}}$, Xiaomi\footnote{https://www.mi.com/prod/xiaomi-ai-glasses}, and RayNeo V3\footnote{https://www.rayneo.com}), spanning 14 domains and 8 categories. 
Constructed through a multi-stage pipeline named \textbf{A}cquirement, \textbf{A}nnotation, \textbf{A}ssessment, and \textbf{A}nalysis (\textbf{\collectorname{}}), \ourname{} incorporates rigorous filtering criteria and human validation to ensure data quality.
The dataset is designed to challenge the agents with the recognition of implicit visual entities from the perspective of smart glasses and the execution of multimodal, multi-hop reasoning. 
Each instance in \ourname{} consists of a domain-specific image representing a specific usage case, along with a question-answer pair and its corresponding retrieval records.
We assess and analyze the quality of our proposed dataset through four aspects: reproducibility, precision, complexity, and correctness, demonstrating its comprehensiveness and suitability for benchmarking multimodal retrieval and reasoning systems in wearable AI applications.

We then investigate the performance of 26 representative VLM agents on the proposed \benchname{} benchmark.
Extensive experiments demonstrate that most of the compared VLMs exhibit suboptimal performance on \ourname{}, with accuracy rates falling below 40\%.
This deficiency is especially pronounced in challenging tasks that require multi-hop reasoning, information-seeking capabilities, and adaptation to rapidly changing scenarios.
Standing out among these models, the large-scale models Gemini 2.5 Pro and GPT-4o, which estimated parameter sizes exceeding 400B, consistently achieve the highest overall performance, reaching accuracies of 43.02\% and 41.91\%, respectively, across all evaluation scenarios.
However, wearable scenarios impose strict constraints on device size and computational resources, necessitating the development of compact models that deliver performance comparable to these large-scale models.
Such models are essential for enabling efficient on-device agents in smart glasses applications.

Building on this insight, we introduce \textbf{\agentname{}}, the first smart glasses agent, designed to deliver accurate answers to multimodal queries by dynamically combining the internal knowledge of VLMs with external information from search engines.
Specifically, \agentname{} is empowered by two key modules: a Demand-Adaptive Answerer and a Dual-Lens Knowledge Retriever.
The former generates direct answers through domain-specific reasoning or calls external tools on demand for retrieval-augmented generation, while the latter automatically integrates object grounding, query decomposition, and multimodal search to supply missing yet relevant knowledge.

Our main contributions are highlighted as follows:
\begin{itemize}[leftmargin=8pt, itemsep=0pt, topsep=1pt, parsep=1pt]
     \item \textbf{\ourname{}: A Comprehensive Benchmark Specifically Tailored for AI Smart Glasses.} 
     We present a comprehensive benchmark specifically designed for smart glasses applications, featuring a reproducible collection process that spans four stages: data acquirement, annotation, assessment, and analysis (\collectorname{}).
     Unlike existing visual question answering datasets, \benchname{} incorporates object grounding, multimodal retrieval, and comprehensive retrieval/tool invocation, establishing it as a challenging and pioneering dataset in the field.
     
    \item \textbf{Leaderboard and Insights.} We benchmark dozens of representative VLMs on \benchname{}.
    The evaluation presents the latest leaderboard of representative multimodal agents on smart glasses-related VQA tasks.
    The corresponding analysis offers insights into the key challenges and future opportunities for smart glasses agents. 
    
    \item \textbf{\agentname{}: The Pioneering Smart Glasses Agent.} We develop \agentname, a demand-adaptive multimodal agent tailored to egocentric and knowledge-intensive reasoning, achieving state-of-the-art performance on \benchname{}.
    By dynamically integrating search engines to access up-to-date external knowledge, the agent outperforms existing VLMs, achieving an average improvement of 2.19\% over the GPT-4o model.
\end{itemize}

\begin{figure*}[t]
\vskip -0.2in
{
\hfill
\begin{minipage}{0.31\textwidth} 
\centering
\captionof{table}{Statistics of \benchname.}
\label{tab:key_statistics}
\resizebox{\textwidth}{!}{
\begin{tabular}{lc}
    \toprule
    \textbf{Statistic} & \textbf{Number} \\
    \midrule
     Total Questions & 2,422 \\
     ~- Multiple-hop Questions &  775 \\
     ~- Single-hop Questions & 1,647 \\
    Total Scenarios & 14 \\
    Total Website Sources & 519 \\
    \midrule
    Total Number of Images & 2,422\\
    ~- Ray-Ban Meta &  257 \\
    ~- Xiao Mi &  797 \\
    ~- RayNeo V3 &  1,368 \\
    Average Image Size (px) & 3,630 $\times$ 2,887 \\
    \midrule
    Maximum Question Length & 51 \\
    Maximum Answer Length & 171 \\
    Average Question Length & 10.18 \\
    Average Answer Length & 18.19\\
    Average Hop Number & 1.39 \\
    Total Tool-Usage Number & 2, 011 \\
    Average Tool-usage Number & 0.83 \\
    \bottomrule
\end{tabular}
}
\end{minipage} 
\hfill
\begin{minipage}{0.5\textwidth}
    \centering
    \captionof{figure}{Data distribution on Difficulty, Domain, and Category.}
    \label{fig:source_dataset}
    \includegraphics[width=\linewidth]{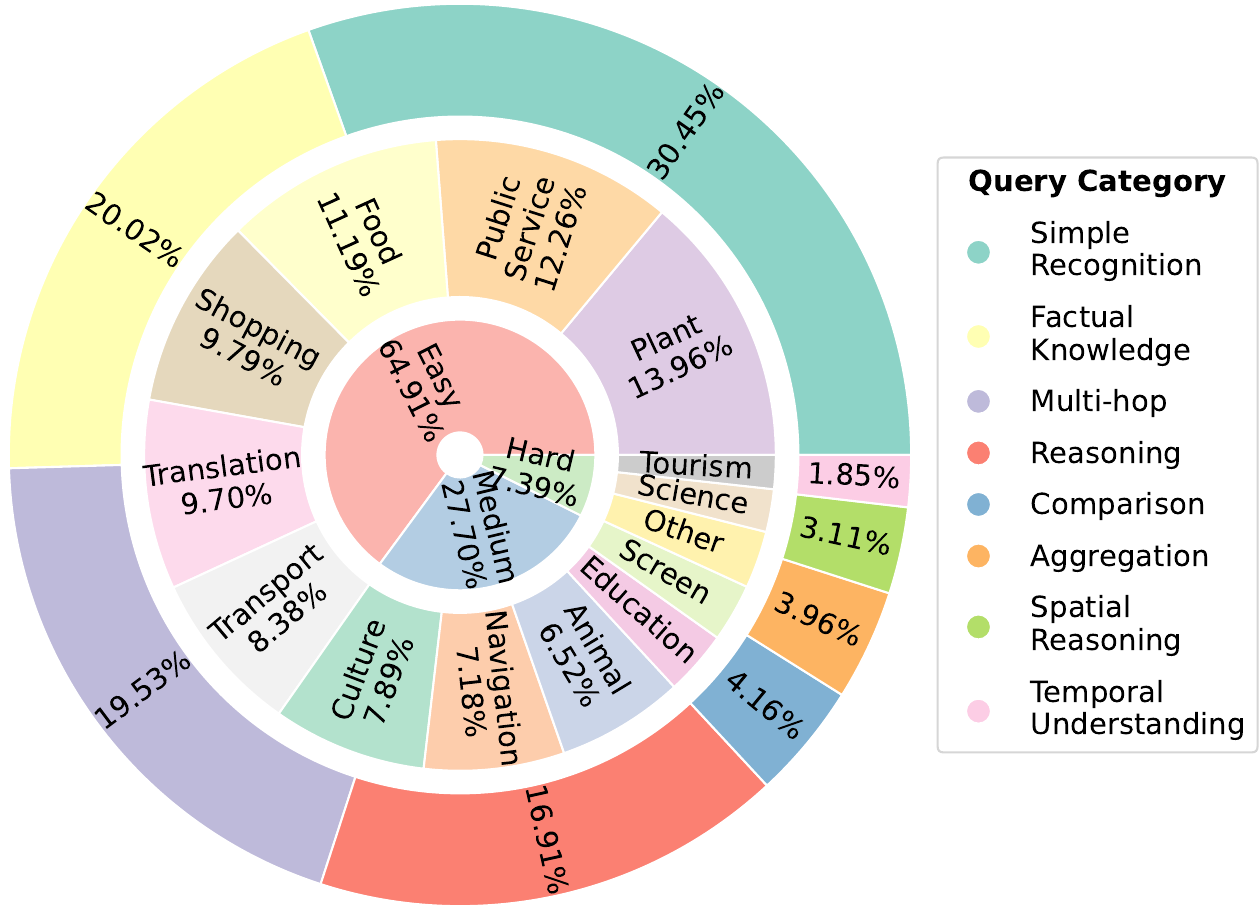}
\end{minipage}
\hfill
}
\vskip -0.1in
\end{figure*}

%% file: sections/2_GlassesBench.tex
\section{\benchname}


\subsection{A$^4$ Data Collection Pipeline}
\label{sec:pipeline}

To construct \benchname, we design a rigorous data collection pipeline termed \textbf{\collectorname{}}, which consists of four tightly integrated stages: \textbf{Acquirement}, \textbf{Annotatation}, \textbf{Assessment}, and \textbf{Analysis}.
This pipeline ensures that our dataset is realistic, comprehensive, and traceable for evaluating VLMs in diverse smart glasses scenarios (Figure~\ref{fig:a4_pipeline}).

\noindent \textbf{Step 1: Acquirement.}
The first stage involves the systematic collection of authentic egocentric data using smart glasses.
Participants equipped with smart glasses devices record diverse real-world scenarios encompassing various environments, objects, and human-centered activities.
This stage yields a comprehensive repository of raw image–query pairs that serves as the foundation for subsequent annotation and reasoning-oriented processing.

\noindent \textbf{Step 2: Annotatation.}
The second stage enriches the raw image–query pairs with structured metadata and retrieval traces.
Human annotators label each sample with key attributes such as category, question type, difficulty, and reasoning steps, while performing both visual and textual search to simulate real-world information seeking.
The focused image regions, decomposed sub-queries, and supporting results are recorded, forming a traceable dataset in which each answer is explicitly grounded in its evidence.

\noindent \textbf{Step 3: Assessment.}
The third stage performs a four-step assessment:
(1) Rule Assessment automatically validates basic constraints such as question length, hop count, and tool usage;
(2) Agent Assessment uses a reference multimodal model to verify query splitting, attribute labels, and answer reproducibility;
(3) Expert Assessment involves human reviewers rating reproducibility, precision, complexity, and correctness; and
(4) Query Filtering aggregates the above results to retain high-quality samples, forming a robust loop that continuously refines data quality.

\noindent \textbf{Step 4: Analysis.}
The final stage analyzes the dataset from a macro perspective, ensuring balanced sample distributions across four key aspects: query and answer lengths, query prefixes, and common topics in queries and answers. These evaluations confirm the dataset’s suitability as a robust benchmark for assessing the multimodal retrieval and reasoning capabilities of smart glasses agents. Details are provided in Appendix~B.

\begin{figure*}[t]
    \vskip -0.3in
    \centering
    \includegraphics[width=0.9\linewidth]{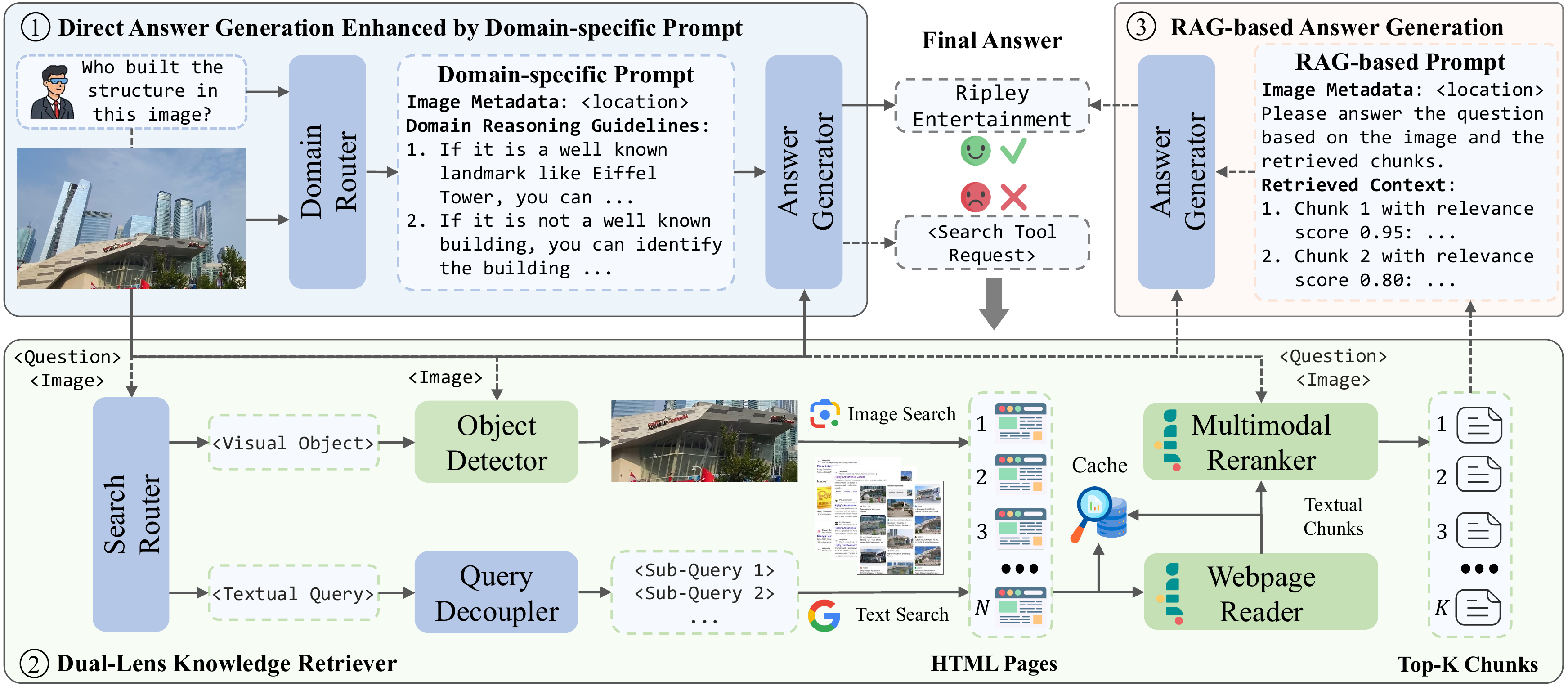}
    \caption{Overview of the proposed \agentname{}, composed of a Demand-Adaptive Answerer and a Dual-Lens Knowledge Retriever. All modules marked in blue are powered by VLMs, while modules in green are external tools.}
    \label{fig:agent}
\end{figure*}

\input{tables/dataset-comparison}

\subsection{Data Description}
\label{sec:description}

\noindent \textbf{Dataset Statistics.} 
As shown in Table~\ref{tab:key_statistics}, \benchname{} contains \querynum{} egocentric image-question pairs collected using three representative commercial smart glasses: \textbf{Ray-Ban Meta}, \textbf{XiaoMi}, and \textbf{RayNeo}, which contribute 257, 797, and 1,368 images, respectively.
As shown in Figure~\ref{fig:source_dataset}, they span 14 practical domains and 8 fine-grained query categories, reflecting a wide range of real-world information needs encountered by smart glasses users.
Each instance includes a high-resolution egocentric image, a natural-language question, and the full interaction search traces, tool usage logs, and step-by-step reasoning paths.
Among all instances, 775 questions involve multi-hop reasoning, while the remaining 1,647 require a single hop.
To underscore the dataset’s complexity and multimodal depth, we report several key statistics: the longest question spans 61 words, the average reasoning step is 1.39, and a total of 2,011 tool usage events are recorded across the corpus.

\vspace{-13pt}
\noindent \textbf{Comparison with Other Datasets.}
As Table \ref{tab:dataset_comparison} shows, early benchmarks~\cite{chang2022webqa, chen2023can} mainly assess a model’s ability to retrieve external textual knowledge for question answering.
However, visual information may be more advantageous or convenient than textual data in many scenarios.
Recent benchmarks~\cite{fu2025livevqa, jiang2025mmsearch, libenchmarking} have sought to address this limitation by incorporating multimodal search, enabling models to access and utilize diverse web information 
for improved performance on VQA tasks.
Nevertheless, these datasets are not collected through smart glasses and neglect detailed search records or tool-usage trajectories, making them less representative of smart glasses usage scenarios and hindering the ability to trace each answer back to its supporting evidence.
Although the CRAG-MM~\cite{wang2025cragmm} and WearVQA~\cite{chang2025wearvqa} benchmarks were collected using smart glasses, they still lack detailed search logs that would enable inspection of intermediate reasoning steps.
In contrast, our \benchname{} corpus captures egocentric smart glasses footage and logs the full tool-use trajectory, enabling spatially grounded, step-by-step reasoning.
Finally, \benchname{} spans 14 domains and 8 query categories, supporting nuanced evaluation of retrieval strategies and smart glasses agents.

%% file: tables/dataset-comparison.tex
\begin{table*}[t]
\vskip 0.2in
\caption{Comparison with VQA and multimodal search benchmarks across multiple dimensions. }
\centering
\small
\resizebox{\textwidth}{!}{
\begin{tabular}{lcccccccccccc}
\toprule
\textbf{Dataset} & \textbf{Language} & \textbf{Hops}  &\textbf{Image Source} &  \textbf{Location} &  \textbf{MM Search}  &  \textbf{Obj. Detect.} & \textbf{Search Log}  & \textbf{\#Domain} & \textbf{\#Category} \\
\midrule
WebQA~\cite{chang2022webqa}  &en & multi-hop & Web & \xmark & \xmark & \xmark &  \xmark & - & 2 \\
InfoSeek~\cite{chen2023can} &en & single-hop & Web   & \xmark & \xmark & \xmark &   \xmark & 9 & - \\
MRAG-Bench~\cite{hu2024mrag}  &en & single-hop  & Web & \xmark  & \cmark & \xmark   & \xmark & 9 & 3 \\
LIVEVQA~\cite{fu2025livevqa}  &en & multi-hop & News & \xmark  & \cmark & \xmark  & \xmark & 12 & - \\
MMSearch~\cite{wu2025mmsearch}  &en & multi-hop  & Web  & \xmark  & \cmark & \xmark  & \xmark & 14 & - \\
Dyn-VQA~\cite{libenchmarking} &en/zh & multi-hop  & Web  & \xmark & \cmark & \xmark   & \xmark & 9 & 3 \\
CRAG-MM~\cite{wang2025cragmm} & en & multi-hop  & Glasses \& Web  & \xmark & \cmark & \cmark   & \xmark & 13 & 4 \\
WearVQA~\cite{chang2025wearvqa} & en & multi-hop  & Glasses \& Web  & \xmark & \xmark & \cmark   & \xmark & 7 & \textbf{10} \\
\midrule
\textbf{\benchname} &en/zh/jp/fr & 1 - 4 hops & Smart Glasses & \cmark & \cmark & \cmark & \cmark & \textbf{14} & 8 \\
\bottomrule
\end{tabular}
}
\label{tab:dataset_comparison}
\vskip -0.15in
\end{table*}

%% file: sections/3_SuperGlass_Agent.tex
\section{\agentname{}}
\vspace{-10pt}
The proposed \benchname{} poses challenges grounded in egocentric, real-world conditions characterized by broad visual scopes, extensive knowledge demands, and multi-hop reasoning, where existing VLMs \citep{liu2024improved,wu2024deepseek,bai2025qwen25vl,zhu2025internvl3} often exhibit limited effectiveness.
To tackle these challenges, we introduce \agentname{}, a strong baseline tailored for smart glasses scenarios. \agentname{} is capable of delivering accurate responses to user queries by adaptively integrating the intrinsic knowledge of VLMs with external evidence retrieved through a well-designed retrieval-augmented framework.
As shown in Figure~\ref{fig:agent}, \agentname{} consists of two core components:
(1) a Demand-Adaptive Answerer, which produces direct answers via domain-specific reasoning or invokes retrieval-augmented generation when additional context is required; and
(2) a Dual-Lens Knowledge Retriever, which gathers and ranks external information from both visual and textual lenses to enrich knowledge coverage and enhance reasoning accuracy.

\input{tables/learderboard}

\subsection{Demand-Adaptive Answerer}
When faced with questions of varying scope and complexity, humans instinctively assess whether their internal knowledge is sufficient or external information is needed~\cite{cocchi2025augmenting}.
Inspired by this behavior, we design a demand-adaptive answerer that dynamically switches between direct response and retrieval-augmented generation.
Given a question–image pair, the Answerer first attempts to answer using internal knowledge acquired from large-scale pretraining corpora.
However, as queries in \benchname{} span diverse domains, such as culture, education, and public services (Figure~\ref{fig:dataset_intro}), uniform processing often leads to suboptimal reasoning due to domain mismatch and limited contextual grounding~\cite{jiang2025qadragon}.
To address this, we instruct the VLM to act as a domain router, identifying the semantic domain of each question–image pair and applying domain-specific prompts aligned with predefined categories.
This routing mechanism activates relevant internal knowledge and enhances domain-aware reasoning.
For rare or long-tailed queries beyond the model’s internal scope, the Answerer triggers a multimodal retrieval pipeline to gather external evidence, which is then integrated into an RAG prompt to synthesize the final answer.
This adaptive design strikes a balance between efficiency and accuracy, enabling \agentname{} to perform robustly in both offline and online settings.

\subsection{Dual-Lens Knowledge Retriever}
As depicted at the bottom of Figure~\ref{fig:agent}, \agentname{} integrates online search engines (e.g., Google) into its retrieval process through two search branches, \emph{i.e.}, image and text, reflecting the inherently multimodal nature of smart glasses QA tasks that rely on both visual and textual knowledge~\cite{lin2022retrieval,yuan2025mkg}.
To select suitable search tools, a VLM-based Search Router decomposes the required knowledge into two modalities: \emph{visual objects} from the input image and \emph{textual queries} from the question.
For visual objects, an open-vocabulary Object Detector~\citep{cheng2024yolo,liu2024grounding} identifies precise regions in the image, as relevant objects often occupy small areas in the field of view of smart glasses.
For textual queries, which often require multi-hop reasoning, a Query Decoupler divides them into single-hop sub-queries to reduce search complexity.
The extracted object regions and sub-queries are then passed to the search engine api for image and text retrieval, returning the top-$N$ raw HTML pages as candidate evidence.

Since raw HTML pages are not VLM-friendly due to excessive length, irregular layout, and redundant tags, we employ a Webpage Reader~\citep{wang2025readerlm} to extract and clean the textual content.
To mitigate latency from search and HTML parsing, we implement a two-layer Redis cache: the first maps text or image queries to webpage URLs, thereby avoiding redundant searches, and the second maps URLs to parsed content, preventing repeated parsing.
A Multimodal Reranker~\citep{jina-reranker} further ensures relevance by segmenting the cleaned content into chunks and reordering them based on weighted similarity to both textual and visual queries. Finally, the top-ranked chunks are integrated into the RAG prompt to guide the Answerer in producing the final response. Please refer to Appendix~D for more details.

\begin{figure*}[t]
\vskip -0.3in
\begin{minipage}{0.31\textwidth}
\centering
    \includegraphics[width=\linewidth]{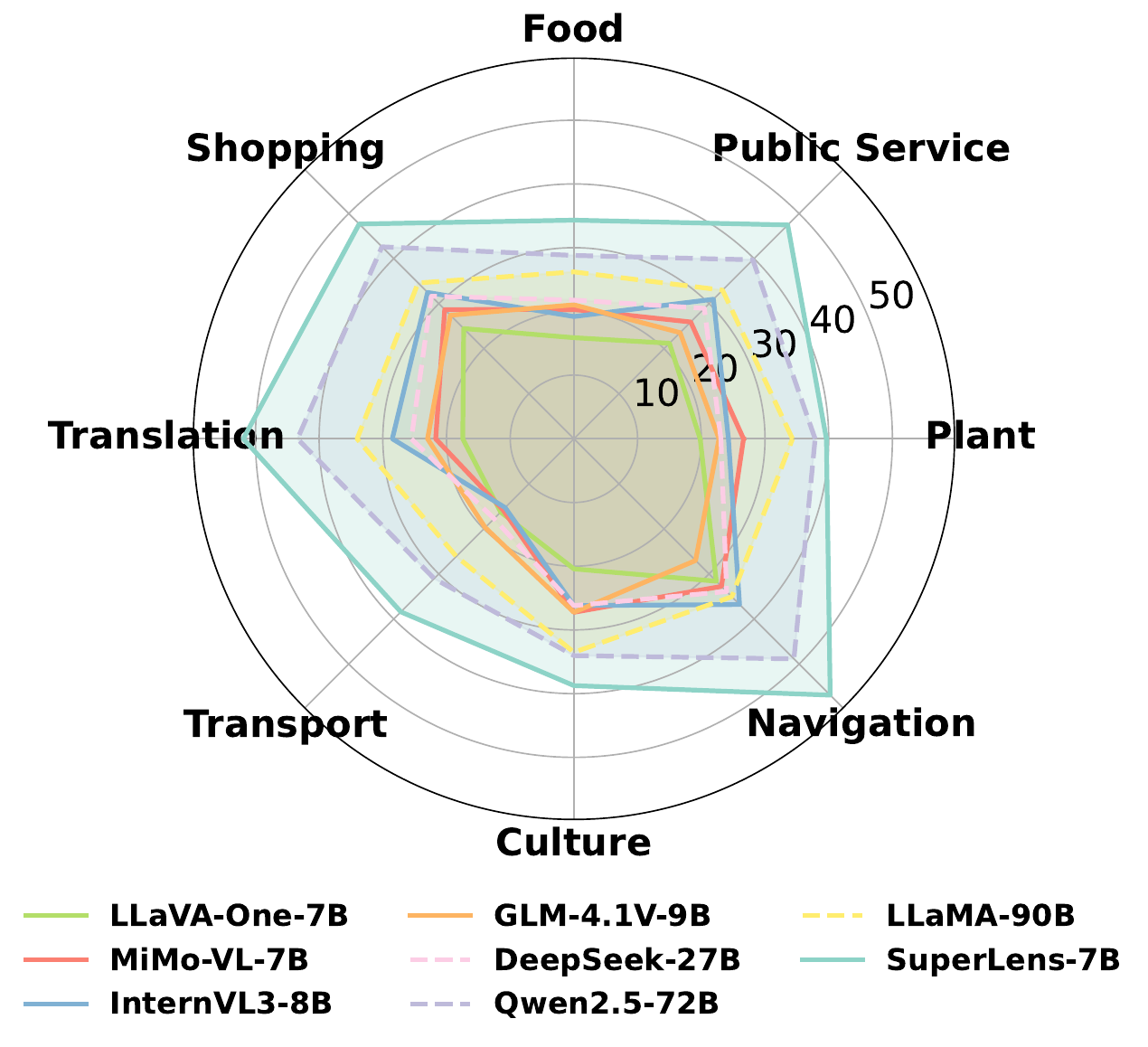}
    \caption{Performance across domains.}
    \label{fig:domain-chart}
\end{minipage}
\hfill
\begin{minipage}{0.31\textwidth}
\centering
    \includegraphics[width=\linewidth]{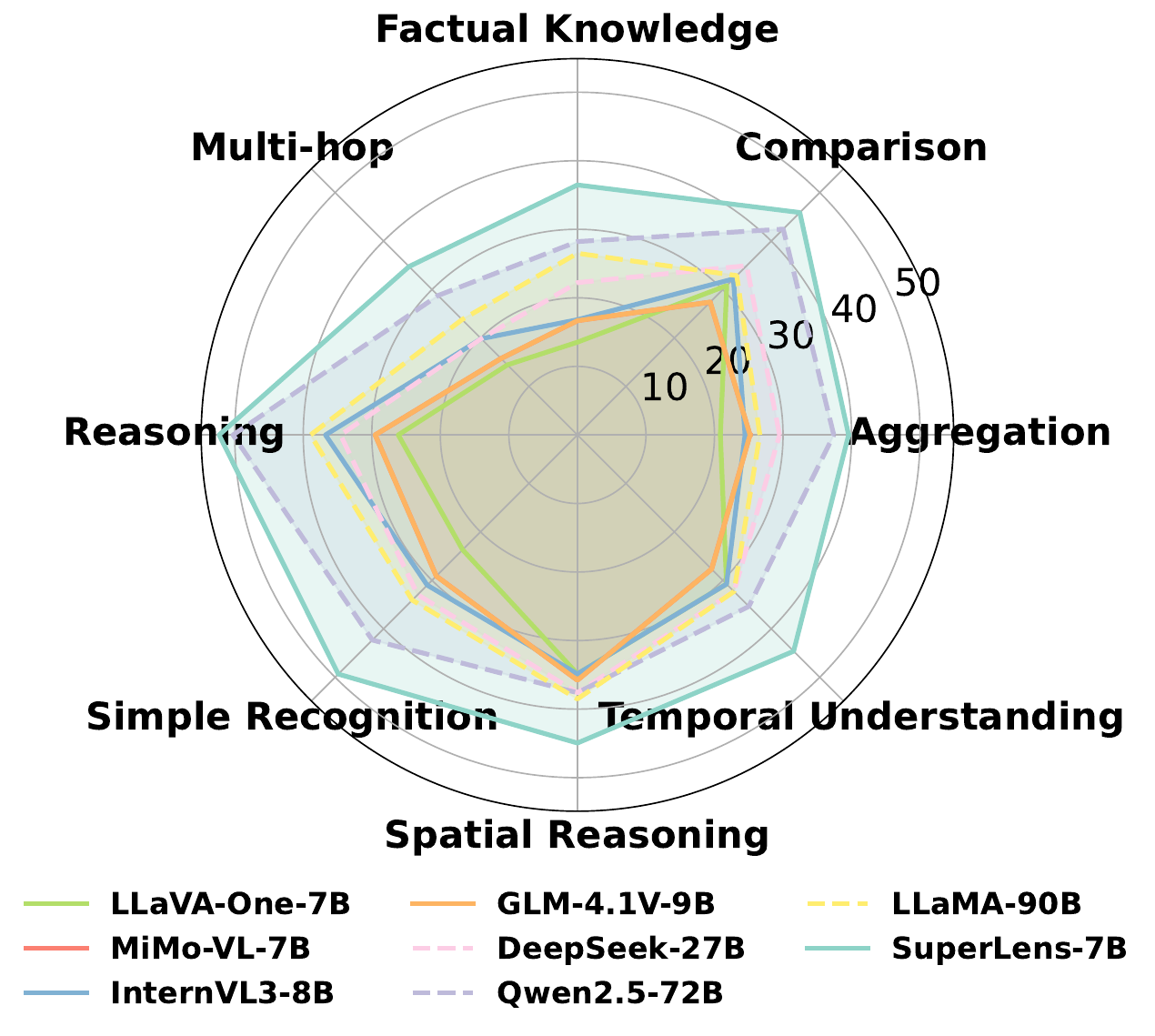}
    \caption{Performance across categories.}
    \label{fig:category-chart}
\end{minipage}
\hfill
\begin{minipage}{0.37\textwidth}
\centering
    \includegraphics[width=\linewidth]{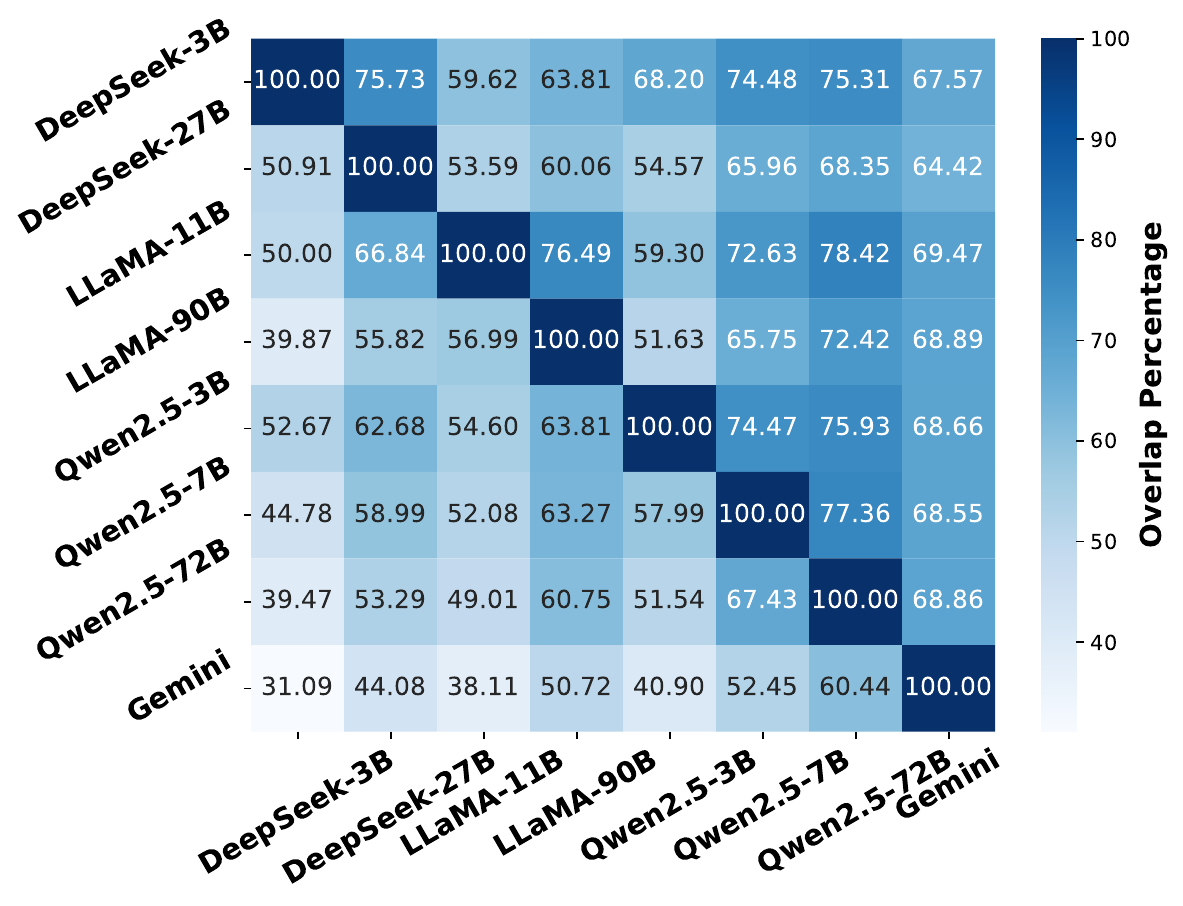}
    \caption{Overlap between correct responses.}
    \label{fig:overlap-heatmap}
\end{minipage}
\vskip 0.2in
\end{figure*}

%% file: tables/learderboard.tex
\begin{table*}[t]
\vskip -0.2in
\caption{\benchname{} Leaderboard: Results of our \agentname{} and 26 Vision Language Models across 3 Dimensions. 
}
\resizebox{\textwidth}{!}{
\begin{tabular}{lcccccccccccc}
\toprule
\multicolumn{1}{c}{\multirow{2}{*}{\textbf{Vision Language Model}}} & \multirow{2}{*}{\textbf{RAG Type}}  & \multicolumn{3}{c}{\textbf{Difficulty}} & \multicolumn{2}{c}{\textbf{Reasoning Steps}} & \multicolumn{4}{c}{\textbf{Information-Seeking}} & \multirow{2}{*}{\textbf{All}} & \multirow{2}{*}{\textbf{Rank}} \\
\cmidrule{3-11} 
                         &    & Easy          & Medium         & Hard     & 1-hop     & $\geq$2-hop    & None     & Image    & Text    & Both      &    &                  \\
\midrule
\rowcolor{gray!20} \multicolumn{13}{c}{\textit{{\raisebox{-0.2\height}{\includegraphics[height=1.1em]{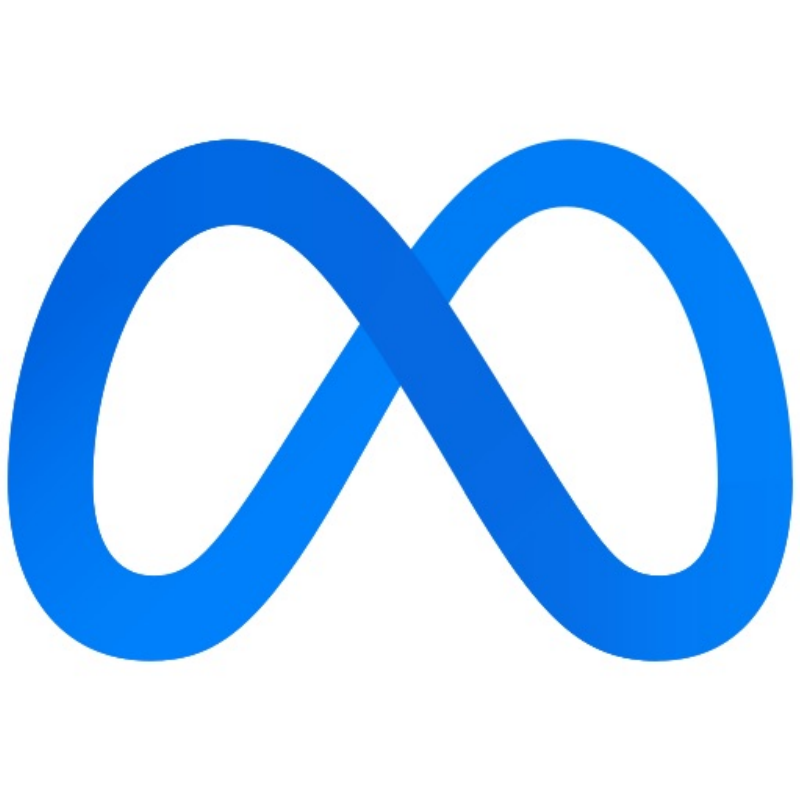}}}~~\textbf{Ray-Ban Meta Smart Glasses}}} \\
\midrule
LLaMA-3.2-11B~\cite{llama-3.2-vision}   & Direct Answer & 27.16 & 17.44 & 14.53 & 26.35 & 17.55 & 33.14 & 13.37 & 19.84 & 14.57 & 23.53 & 15\\
LLaMA-3.2-90B~\cite{llama-3.2-vision}   & Direct Answer & 34.99 & 25.93 & 22.91 & 34.55 & 25.29 & 39.9  & 23.39 & 28.78 & 22.61 & 31.59 & 9\\

\midrule
\rowcolor{gray!15} \multicolumn{13}{c}{{\raisebox{-0.2\height}{\includegraphics[height=1.3em]{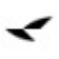}}}~\textit{\textbf{RayNeo Smart Glasses}}} \\
\midrule
Qwen2.5-VL-3B~\cite{bai2025qwen25vl}    & Direct Answer & 30.83 & 20.68 & 18.53 & 29.03 & 21.68 & 36.37 & 17.22 & 21.63 & 12.06 & 25.56 & 12\\
Qwen2.5-VL-7B~\cite{bai2025qwen25vl}    & Direct Answer & 37.72 & 24.59 & 20.67 & 36.92 & 24.13 & 45.59 & 25.71 & 24.23 & 20.35 & 32.82 & 8\\
Qwen2.5-VL-32B~\cite{bai2025qwen25vl}   & Direct Answer & 40.52 & 30.40 & 24.58 & 40.50 & 28.13 & 49.41 & 24.16 & 32.20 & 22.36 & 36.54 & 7\\
Qwen2.5-VL-72B~\cite{bai2025qwen25vl}   & Direct Answer & 41.54 & 31.89 & 25.14 & 41.29 & 29.94 & 50.10 & 24.94 & 33.50 & 24.62 & 37.65 & 4\\
\midrule
\rowcolor{gray!15} \multicolumn{13}{c}{{\raisebox{-0.2\height}{\includegraphics[height=1.1em]{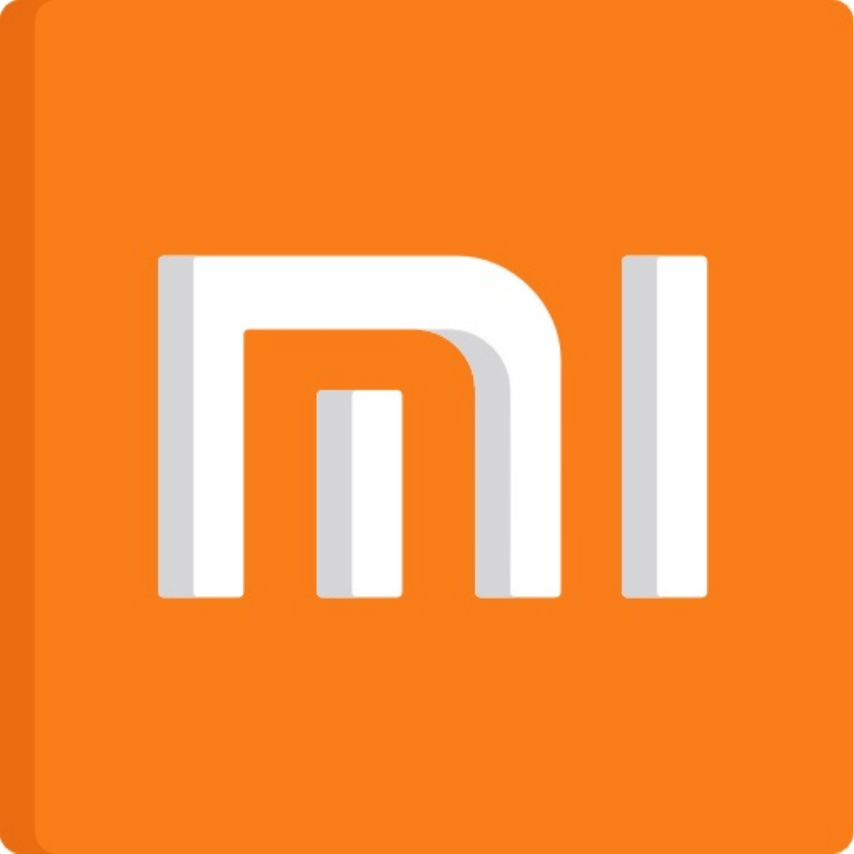}}}~~\textit{\textbf{XiaoMi Smart Glasses}}} \\
\midrule
MiMo-VL-7B~\cite{coreteam2025mimovl}    & Direct Answer & 29.71 & 18.03 & 10.06 & 28.96 & 16.65 & 36.08 & 16.97 & 20.00 & 12.31 & 25.02 & 12\\
\midrule
\rowcolor{gray!15} \multicolumn{13}{c}{{\raisebox{-0.4\height}{\includegraphics[height=1.1em]{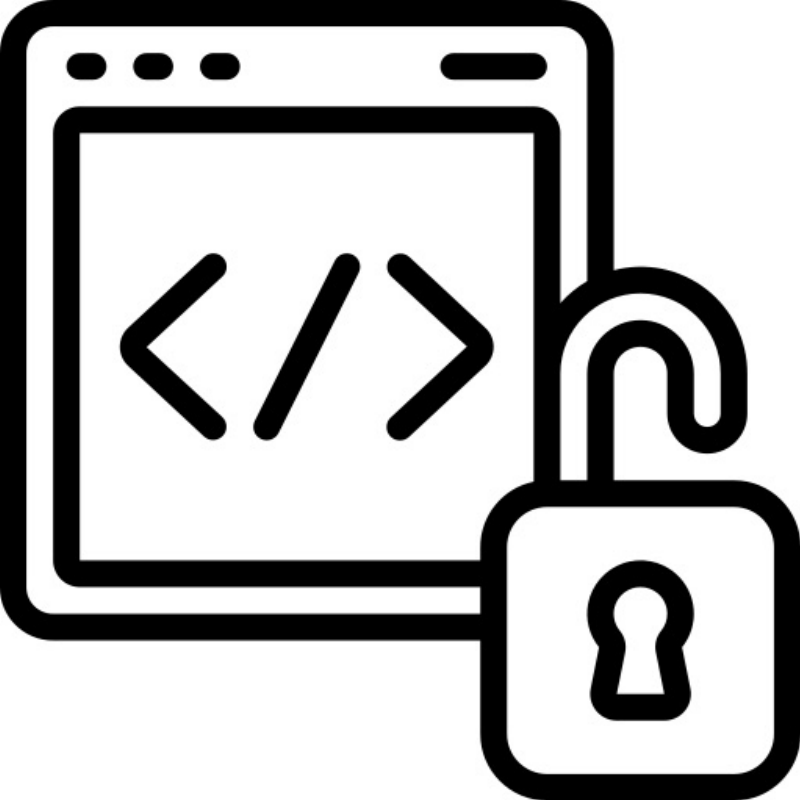}}}~~\textit{\textbf{Open-sourced VLMs}}} \\
\midrule
Phi-3-Vision-4B~\cite{marah2024phi3}    & Direct Answer & 20.74 & 11.18 & 10.06 & 19.98 & 11.61 & 26.18 & 7.20  & 13.98 & 9.55  & 17.30 & 19\\
InternVL3-8B~\cite{zhu2025internvl3}    & Direct Answer & 31.17 & 19.23 & 17.88 & 29.99 & 20.26 & 40.20 & 12.08 & 21.95 & 14.82 & 26.88 & 10\\
GLM-4.1V-9B~\cite{hong2025glm}          & Direct Answer & 27.35 & 17.88 & 15.08 & 26.53 & 18.06 & 33.82 & 14.14 & 19.02 & 15.08 & 23.82 & 14\\
LLaVA-v1.5-7B~\cite{liu2024improved}    & Direct Answer & 12.40 & 6.71  & 7.26  & 11.35 & 8.52  & 15.00 & 4.63  & 9.76  & 5.53  & 10.45 & 27\\
LLaVA-v1.5-13B~\cite{liu2024improved}   & Direct Answer & 13.93 & 7.15  & 9.50  & 13.36 & 8.26  & 16.96 & 4.88  & 10.89 & 6.28  & 11.73 & 26\\
LLaVA-Onevision-0.5B~\cite{li2024llava} & Direct Answer & 16.79 & 8.49  & 6.15  & 15.79 & 9.29  & 21.08 & 8.23  & 10.41 & 5.28  & 13.71 & 24\\
LLaVA-Onevision-7B~\cite{li2024llava}   & Direct Answer & 25.00 & 13.41 & 13.97 & 23.92 & 14.71 & 31.27 & 10.54 & 17.24 & 10.55 & 20.97 & 16\\
DeepSeek-VL2-3B~\cite{wu2024deepseek}   & Direct Answer & 24.11 & 12.22 & 9.50  & 23.13 & 12.52 & 30.78 & 10.80 & 14.15 & 8.79  & 19.74 & 17\\
DeepSeek-VL2-16B~\cite{wu2024deepseek}  & Direct Answer & 28.75 & 16.10 & 10.61 & 27.5  & 16.26 & 35.88 & 15.42 & 16.75 & 12.56 & 23.91 & 13\\
DeepSeek-VL2-27B~\cite{wu2024deepseek}  & Direct Answer & 29.90 & 19.08 & 12.85 & 28.48 & 19.61 & 37.25 & 14.40 & 19.84 & 15.83 & 25.64 & 11\\
\midrule
\rowcolor{gray!15} \multicolumn{13}{c}{{\raisebox{-0.2\height}{\includegraphics[height=1.1em]{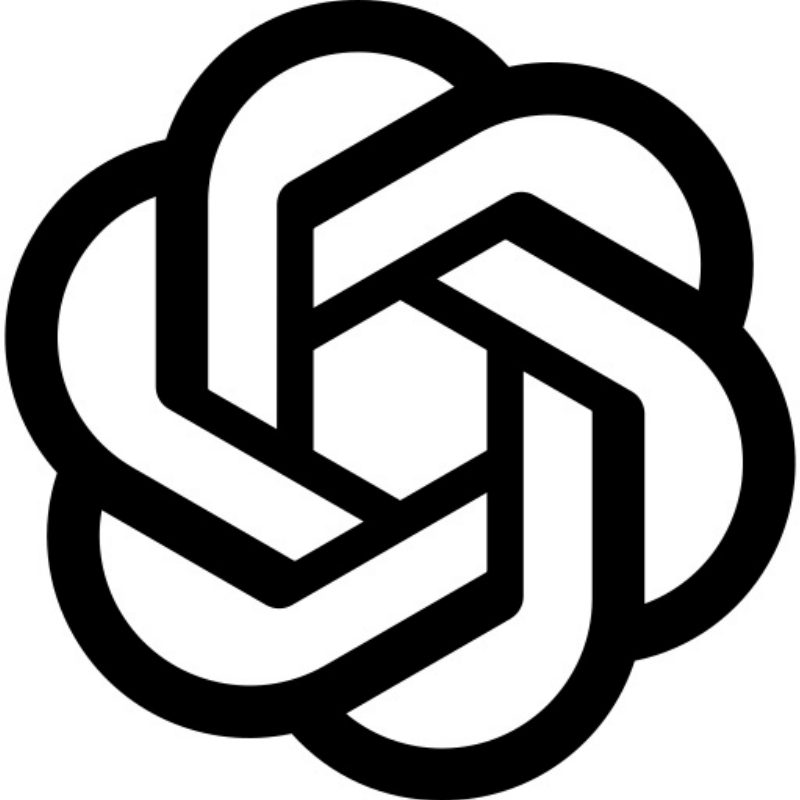}}}~~\textit{\textbf{Proprietary VLMs}}} \\
\midrule
GPT-4o~\cite{achiam2023gpt}           & Direct Answer        & 45.10  & 36.81 & 32.96 & 44.87 & 35.61 & 49.90  & 34.19 & 38.37 & 34.42 & 41.91  & 3\\
Claude 4 Sonnet  & Direct Answer        & 40.14 & 32.34 & 29.05 & 40.32 & 30.45 & 48.63 & 22.88 & 35.12 & 24.87 & 37.16 & 6 \\
Gemini 2.5 Pro   & Direct Answer        & 45.10  & \textbf{39.94} & \textbf{36.31} & 45.36 & \textbf{38.06} & 48.92 & \textbf{37.02} & \textbf{41.95} & \textbf{35.43} & 43.02 & 2 \\
\midrule
\rowcolor{gray!15} \multicolumn{13}{c}{{\raisebox{-0.2\height}{\includegraphics[height=1.1em]{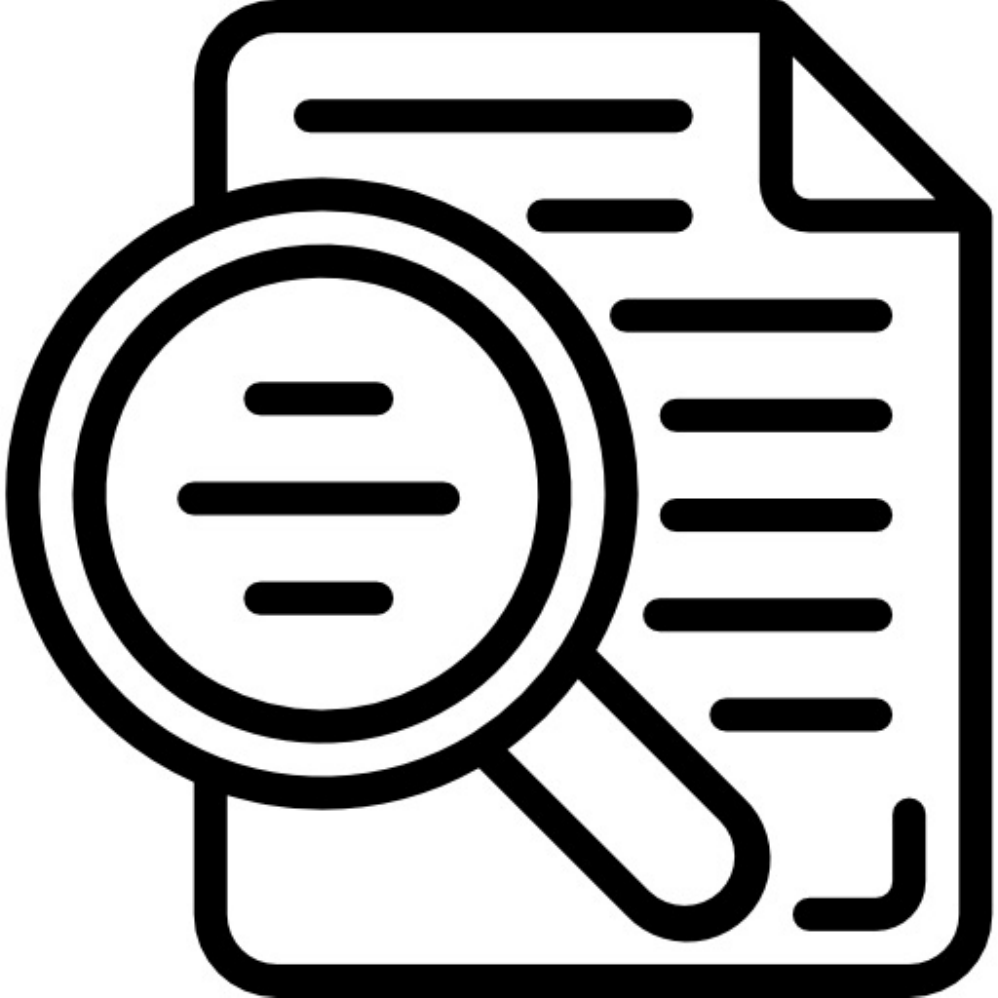}}}~~\textit{\textbf{Heuristic RAG}}} \\
\midrule
LLaMA-3.2-11B   & Image RAG             & 19.97 & 12.97 &  7.82 & 19.06 & 13.03 & 24.41 &  9.00 & 14.80 & 10.05  & 17.13 & 21\\
LLaMA-3.2-11B   & Text RAG              & 15.90 &  8.79 & 10.61 & 15.30 &  9.81 & 16.57 &  6.43 & 17.07 &  7.29  & 13.54 & 25\\
LLaMA-3.2-11B   & Multimodal RAG        & 17.56 &  9.39 & 10.61 & 17.06 &  9.94 & 19.51 &  6.94 & 16.59 &  7.54  & 14.78 & 23\\
Qwen2.5-VL-7B   & Image RAG             & 22.58 & 14.01 & 12.29 & 22.71 & 12.52 & 28.73 & 11.83 & 14.80 & 10.30  & 19.45 & 18\\
Qwen2.5-VL-7B   & Text RAG              & 19.78 & 10.28 &  9.50 & 19.25 & 10.32 & 20.59 & 10.80 & 18.54 &  7.79  & 16.39 & 22\\
Qwen2.5-VL-7B   & Multimodal RAG        & 21.12 &  9.69 & 10.61 & 20.64 &  9.81 & 23.43 & 10.03 & 17.56 &  7.54  & 17.18 & 20 \\
\midrule

\textbf{\agentname{}}\dag~(Ours) &  Multimodal RAG    
& \gaincell{40.84}{13.68}
& \gaincell{31.74}{14.30}
& \gaincell{26.26}{11.73}
& \gaincell{40.01}{13.66}
& \gaincell{31.35}{13.80}
& \gaincell{42.65}{9.51}
& \gaincell{29.05}{15.68}
& \gaincell{39.19}{19.35}
& \gaincell{28.39}{13.82}
& \gaincell{37.24}{13.71}
& 5
\\
\midrule
\textbf{\agentname{}}\ddag~(Ours) &  Multimodal RAG    
& \gaincell{\textbf{49.68}}{11.96}
& \gaincell{34.72}{10.13}
& \gaincell{30.17}{9.50}
& \gaincell{\textbf{48.76}}{11.84}
& \gaincell{34.19}{10.06}
& \gaincell{\textbf{55.78}}{10.19}
& \gaincell{34.70}{8.99}
& \gaincell{40.33}{16.10}
& \gaincell{29.15}{8.80}
& \gaincell{\textbf{44.10}}{11.28}
& 1
\\
\bottomrule
\end{tabular}

}
\raggedright\footnotesize{$^*$ Our smart glasses agent builds upon two base VLMs, Llama-3.2-11B-Vision and Qwen2.5-VL-7B, denoted as \textbf{\agentname{}\dag{}} and \textbf{\agentname{}\ddag{}}. Performance gains in \textcolor{ForestGreen}{green} are measured against their respective backbones.}
\label{tab:leaderboard}
\vskip -0.1in
\end{table*}

%% file: sections/4_Experiment.tex
\section{Experiment}
\subsection{Experiment Setup}
\textbf{Evaluation Models.} 
The dataset is collected using three representative commercial and various smart glasses platforms: Ray-Ban Meta, XiaoMi, and RayNeo. 
Specifically, we examine models from the \emph{LLaMA-3.2-Vision}~\citep{llama-3.2-vision}, \emph{MiMo-VL}~\citep{coreteam2025mimovl}, and \emph{Qwen2.5-VL}~\citep{bai2025qwen25vl} families.
Beyond these, we include a diverse set of mainstream open-source VLMs, such as \emph{Phi-3-Vision}~\citep{marah2024phi3}, \emph{InternVL3}~\citep{zhu2025internvl3}, \emph{GLM-4.1V}~\citep{hong2025glm}, \emph{LLaVA-v1.5}~\citep{liu2024improved}, \emph{LLaVA-OneVision}~\citep{li2024llava}, and \emph{DeepSeek-VL2}~\citep{wu2024deepseek}, as well as proprietary models including \emph{GPT-4o}, \emph{Claude 4 Sonnet}, and \emph{Gemini 2.5 Pro}.
To further assess the impact of explicit retrieval augmentation in smart glasses contexts, we develop a set of heuristic RAG variants by enhancing LLaMA-3.2-11B and Qwen2.5-VL-7B with three retrieval strategies: image-only, text-only, and multimodal RAG.
In total, our evaluation encompasses 26 vision language models, comprising 20 foundation models and 6 RAG-enhanced variants.

\noindent \textbf{Implementation Details.}
To accommodate high-resolution input images, we resize their shortest edge to 1024 pixels for all VLMs, preventing out-of-memory errors.
In our search pipeline, both image and text queries retrieve up to 5 relevant webpages. The Multimodal Reranker then selects the top 10 ranked chunks, while discarding chunks with scores below 0.6. 
Similarity weights are set to 0.6 for questions and 0.4 for images. Refer to Appendix~D for more details.

\noindent \textbf{Evaluation Metrics.} 
The LLM-as-Judge framework~\cite{wu2025mmsearch} is utilized as our evaluation metric. Specifically, we provide the original question, the ground-truth answer, and the model-generated response to Qwen2.5-32B~\citep{qwen2025qwen25technicalreport}, which evaluates whether the response accurately captures all key information from the ground truth. The complete evaluation prompt is provided in Appendix~E.1.

\subsection{Leaderboard}
Table~\ref{tab:leaderboard} reports the performance of our smart glasses agent \agentname{}, alongside 26 leading VLMs (20 in the direct-answer setting and 6 with heuristic RAG) on the proposed \benchname{} dataset.
From the results, we can make the following observations.
\textbf{(1) The proposed \benchname{} poses a formidable challenge to all open-source, proprietary, and RAG-based VLMs}, as even the most advanced model (i.e., Gemini 2.5 Pro) achieves only around 43\% accuracy.
Across difficulty levels, all models exhibit clear performance declines from Easy to Hard questions (e.g., GPT-4o drops from 45.10\% to 32.96\%).
Moreover, for queries requiring multi-hop reasoning, external knowledge seeking, or fast-changing information (see Appendix~F), all models consistently underperform compared to other queries within the same taxonomy.
These results highlight that current VLMs struggle to effectively handle complex user queries in smart glasses scenarios.
\textbf{(2) Open-source VLMs still trail their closed-source counterparts by a significant margin.}
GPT-4o and Gemini 2.5 Pro outperform all open-source models, achieving overall accuracies of 41.91\% and 43.02\%, respectively. 
Although Qwen2.5-VL-72B slightly surpasses Claude 4 Sonnet, it still lags behind Gemini 2.5 Pro by 6.45\%. 
These results indicate substantial room for advancing open-source smart glasses agents.
\textbf{(3) VLMs, especially those within the same family, such as Qwen and DeepSeek, follow a clear scaling law}, with performance improving markedly as model size increases.
This trend is particularly evident in tasks involving multi-hop reasoning and the retrieval of rare knowledge.
\textbf{(4) Naively applying heuristic RAG strategies does not yield overall performance gains.} As shown in \textit{Heuristic RAG} part of Table~\ref{tab:leaderboard}, ineffective retrieval often disrupts the generation process, underscoring the importance of appropriate multimodal RAG design for complex smart glasses scenarios.

Our proposed \textbf{\agentname{}\ddag}, which integrates a demand-adaptive answerer with a dual-lens knowledge retriever, achieves an overall score of \textbf{44.10\%}, the highest among all VLMs in Table~\ref{tab:leaderboard}.
Compared to the baseline Qwen2.5-VL-7B, it attains a notable \textbf{11.28\%} improvement, highlighting the precision and effectiveness of our multimodal RAG strategy for complex, knowledge-intensive user queries.
Figures~\ref{fig:domain-chart} and~\ref{fig:category-chart} provide a detailed breakdown across image domains and query categories, where \agentname{} consistently outperforms all open-source models, with pronounced advantages in real-world scenarios such as \textit{Public Service} and \textit{Navigation}, as well as in cognitively demanding categories like \textit{Factual Knowledge} and \textit{Spatial Reasoning}. More analyses can be found in Appendix~F.
Despite our progress, \agentname{} represents only an initial step toward smart glasses agents, as its performance remains below 45\% on \benchname{}. 
\textbf{Developing stronger VLM-based agents for smart glasses thus remains a promising direction for future research.}

\subsection{Ablation study on \textbf{\agentname{}}}
\input{tables/big_ablation}
As shown in Table~\ref{tab:ablation}, the performance of \agentname{} stems from the tight coordination between the Demand-Adaptive Answerer and the Dual-Lens Knowledge Retriever.
Under the Demand-Adaptive Retrieval setting, ablating retriever components highlights their respective contributions: removing the Query Decoupler causes a 3.18\% drop, while disabling the Object Detector leads to a smaller 1.53\% decline.
Notably, using image-only search results in a 1.53\% drop, much larger than the 0.29\% decline from text-only search, suggesting that our \agentname{} relies more on accurate textual retrieval.
The importance of the Demand-Adaptive Answerer is further demonstrated in the Forced Retrieval setting: without its control, indiscriminate retrieval severely degrades performance (-27.71\% for text-only and -24.65\% for image-only), showing that incorrect retrieval can be worse than no retrieval at all.
While removing either retrieval branch individually causes only mild degradation, removing both leads to a sharp 11.28\% drop, underscoring the complementarity of dual-lens retrieval.

\subsection{Experimental Analysis}
\noindent \textbf{Appropriate Retrieval Strategy.}
Table~\ref{tab:leaderboard} demonstrates that ineffective retrieval can substantially degrade model performance, sometimes underperforming direct-answer baselines.
For instance, LLaMA-3.2-11B reaches 23.53\% without retrieval, whereas its heuristic RAG variants decrease to 17.13\% (Image RAG), 13.54\% (Text RAG), and 14.78\% (Multimodal RAG).
This decline arises from the complexity of \benchname{}, which often requires detecting non-vision-centric objects and handling multi-hop queries. Heuristic RAG, relying on direct image or text retrieval, fails to capture these fine-grained search needs and instead introduces substantial noise.
By contrast, \agentname{} combines demand-adaptive retrieval control with a dual-lens knowledge retriever to obtain high-quality evidence context, outperforming heuristic RAG by 24.65\%.

\noindent \textbf{Textual vs. Visual Retrieval.}
In Table~\ref{tab:ablation}, text-only retrieval surpasses its image-only counterpart by 1.24\%.
We attribute this advantage to the fact that most knowledge is encoded in textual form, making textual retrieval more effective for accessing relevant information. 
These findings highlight the pivotal role of textual retrieval in RAG and emphasize the need for precise, relevance-aware visual retrieval to enable stronger multimodal RAG.

\noindent \textbf{Varying Prediction Behaviors.}
We further analyze the overlap of correctly answered questions across different models.
From Figure~\ref{fig:overlap-heatmap}, we can see that: 
(1) Although Gemini 2.5 Pro achieves strong overall scores, its agreement with open-source models remains below 70\%, suggesting distinct knowledge and capability distributions.
(2) High-score models do not simply excel by answering harder questions while also covering the easy ones solved by weaker models, consistent with the findings of Dyn-VQA~\cite{libenchmarking}.
(3) Models from the same family exhibit higher overlap, likely due to shared training data and architectural paradigms.

\begin{figure}
\vskip -0.3in
    \centering
    \includegraphics[width=\linewidth]{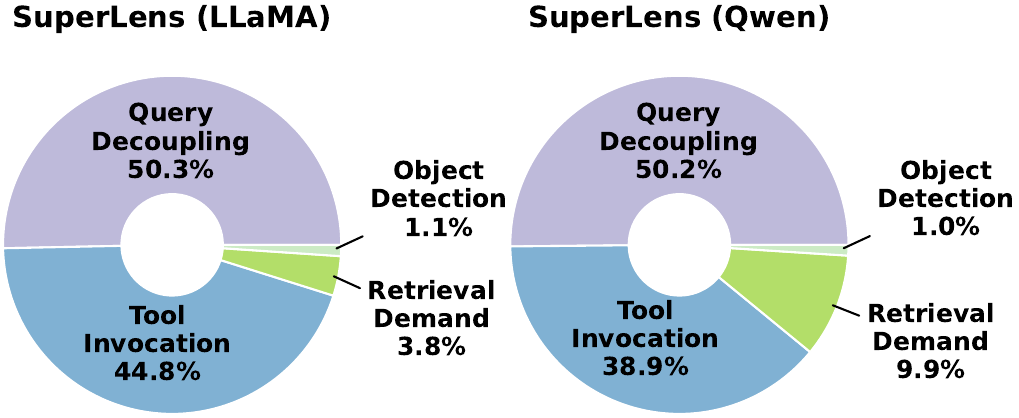}
    \caption{Distribution of error types.}
    \label{fig:error_analysis}
    \vskip 0.1in
\end{figure}

\noindent \textbf{Error Analysis.}
We present a comprehensive analysis of the error types observed in \agentname{} by comparing its intermediate responses with the annotated search logs in \benchname{}.
As illustrated in Figure~\ref{fig:error_analysis}, both \agentname{} variants are primarily affected by query decoupling and tool invocation errors, highlighting current limitations in understanding complex queries and effectively utilizing external tools, a direction we leave for future work.

%% file: tables/big_ablation.tex
\begin{table}[t]
\vskip -0.2in
\centering
\caption{Ablation on key designs of our \agentname{}. }
\label{tab:ablation}
\resizebox{\linewidth}{!}{
\begin{tabular}{lcc}
\toprule
\textbf{Setting} & \textbf{All} & \textbf{Decline} \\
\midrule
\textbf{\agentname{}} & \textbf{44.10} & -- \\
\midrule
\multicolumn{3}{l}{\textbf{A. No Retrieval}} \\
\quad w/o Search                                                     & 32.82 & \declinecell{11.28} \\
\midrule
\multicolumn{3}{l}{\textbf{B. Mandatory Retrieval }} \\
\quad Image Only (w/o Object Dectector)                              & 19.45 & \declinecell{24.65} \\
\quad Text Only (w/o Query Decoupler)                                & 16.39 & \declinecell{27.71} \\
\quad Image + Text (w Object Dectector + w Query Decoupler)          & 32.04 & \declinecell{12.06} \\
\quad Image + Text (w/o Object Dectector + w/o Query Decoupler)      & 16.45 & \declinecell{27.65} \\
\midrule
\multicolumn{3}{l}{\textbf{C. Demand-Adaptive Retrieval}} \\
\quad Image Only (w Object Dectector)                                & 42.57 & \declinecell{1.53} \\
\quad Image Only (w/o Object Dectector)                              & 36.58 & \declinecell{7.52} \\
\quad Text Only (w Query Decoupler)                                  & 43.81 & \declinecell{0.29} \\
\quad Text Only (w/o Query Decoupler)                                & 38.03 & \declinecell{6.07} \\
\quad Image + Text (w Object Dectector + w/o Query Decoupler)        & 40.92 & \declinecell{3.18} \\
\quad Image + Text (w/o Object Dectector + w Query Decoupler)        & 43.44 & \declinecell{0.66} \\
\bottomrule
\end{tabular}
}
\end{table}

%% file: sections/5_Conclusion.tex
\section{Conclusion}
In this work, we propose \textbf{\benchname}, the first smart glasses VQA benchmark, comprising \querynum{} real-world image–question pairs across 14 domains and 8 categories.
Our comprehensive evaluation across 26 representative VLMs reveals substantial performance gaps between existing agents and real-world smart glasses usage scenarios.
To address these limitations, we introduce the \textbf{\agentname}, a dedicated system that integrates a Demand-Adaptive Answerer and a Dual-Lens Knowledge Retriever. The \agentname{} significantly outperforms existing models, including a 2.19\% gain over GPT-4o, underscoring the importance of task-aligned designs for smart glasses applications. 

%% file: sections/6_Acknowledgement.tex
\section{Acknowledgments}
The research described in this paper has been partially supported by the General Research Funds from the Hong Kong Research Grants Council (project No. PolyU 15207322, 15200023, 15206024, and 15224524), Hong Kong Research Grants Council’s Theme-based Research Scheme (No. T43-513/23-N), Hong Kong Research Grants Council’s Research Impact Fund (No. R1015-23), Hong Kong Research Grants Council’s Collaborative Research Fund (No. C1043-24GF), Internal research funds from Hong Kong Polytechnic University (project no. P0059586, P0042693, P0048625, and P0051361), and Sheertek International (HK) Limited. This work was supported by computational resources provided by The Centre for Large AI Models (CLAIM) of The Hong Kong Polytechnic University.

%% file: sections/9_Appendix.tex
\newpage

We have included supplementary material to facilitate a more comprehensive understanding and in-depth analysis of the primary paper. The supplementary material is organized as follows:

\begin{itemize}
    \item \textbf{Section~\ref{sec:related_work}: Related Works}
    \item \textbf{Section~\ref{sec:app_dataset}: Details of Dataset Collection}
    \item \textbf{Section~\ref{sec:app_superglasses}: Details of \ourname{}}
    \item \textbf{Section~\ref{sec:app_superlens}: Details of \agentname{}}
    \item \textbf{Section~\ref{sec:app_evaluation}: Details of Evaluation}
    \item \textbf{Section~\ref{sec:app_experiment}: More Experimental Results}
\end{itemize}

\input{sections/1_RelatedWork}

\begin{figure*}[h]
    \centering
    \begin{subfigure}[b]{0.59\textwidth}
        \centering
        \includegraphics[width=\textwidth]{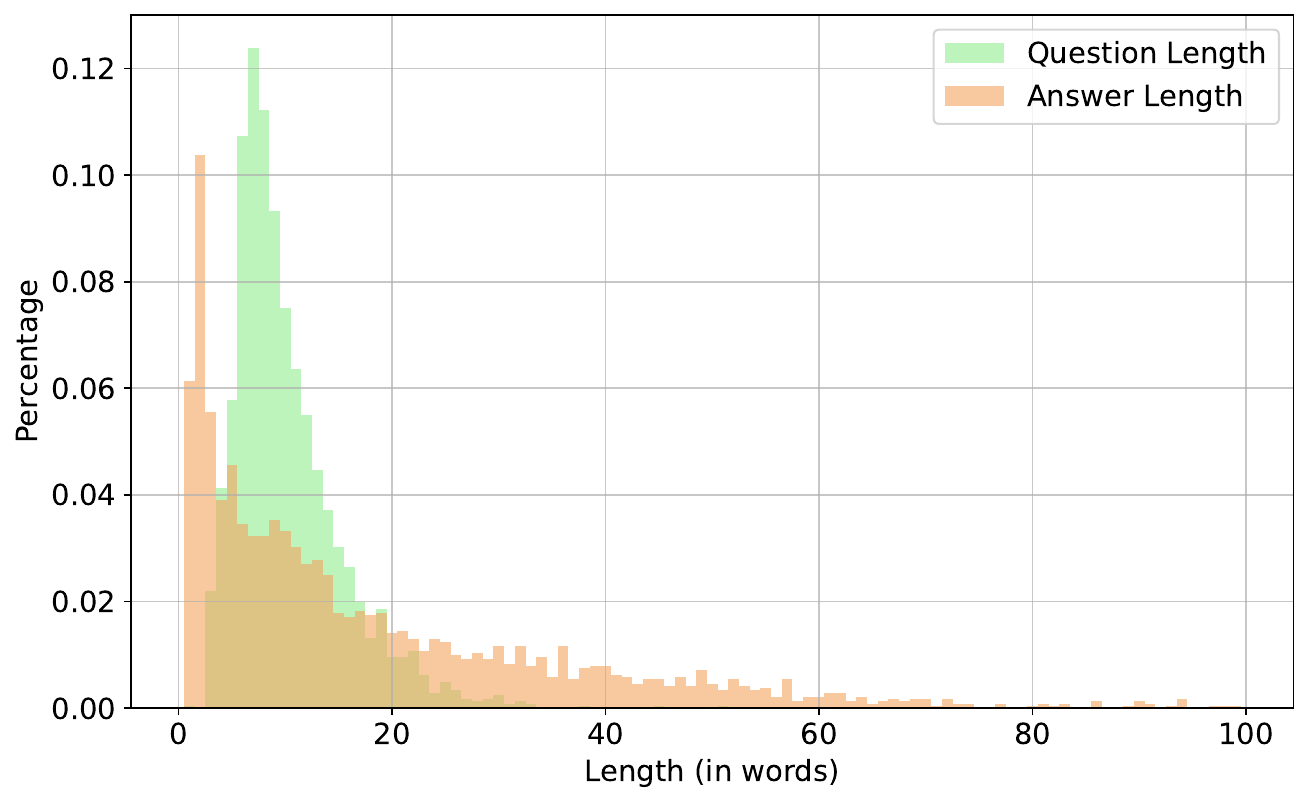}
        \caption{Length distribution of questions and answers.}
        \label{fig:length-distribution}
    \end{subfigure}
    \hfill
    \centering
    \begin{subfigure}[b]{0.39\textwidth}
        \centering
        \includegraphics[width=\textwidth]{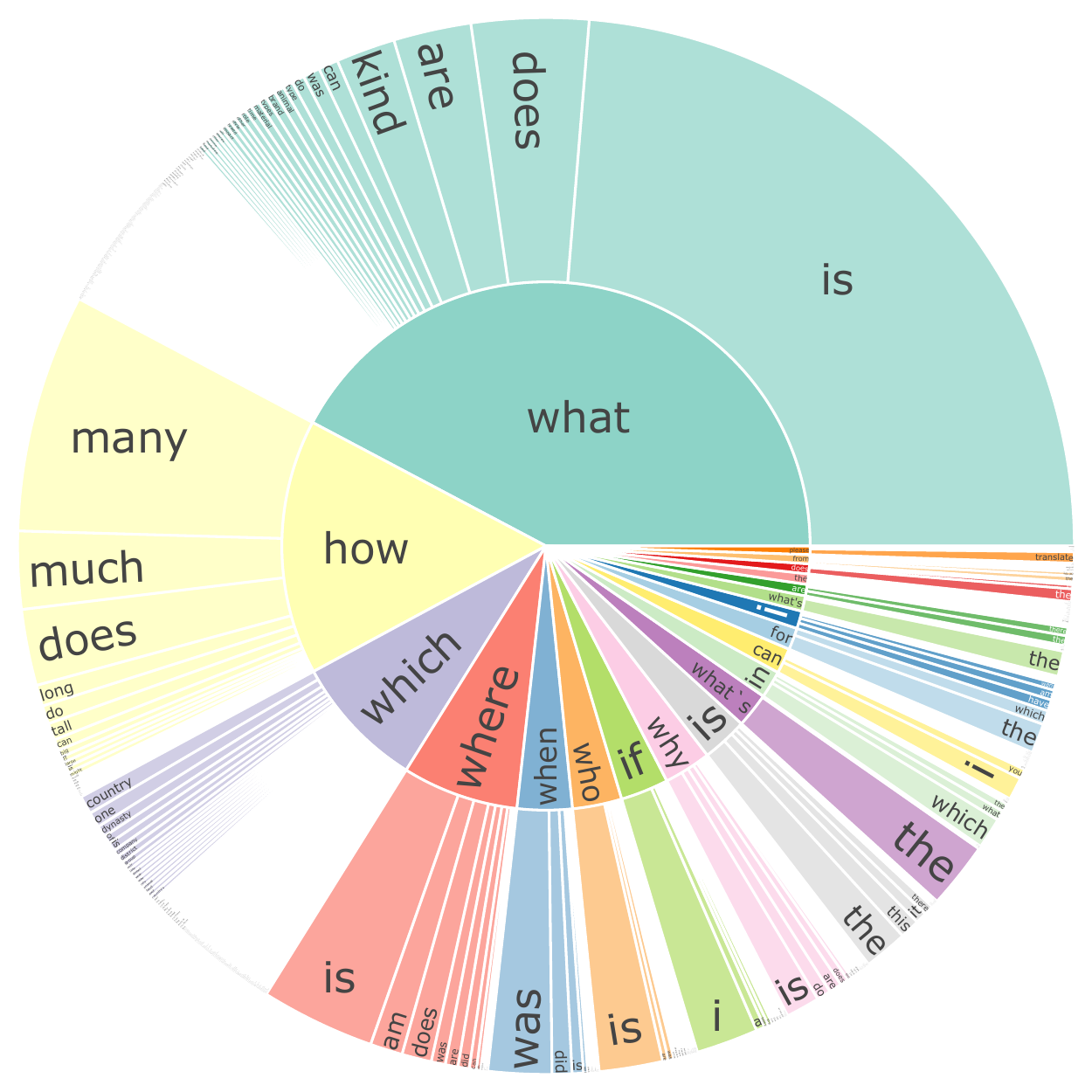}
        \caption{Common question prefixes.}
        \label{fig:question-sunburst}
    \end{subfigure}
    \vspace{10pt}
    \caption{Question/answer length distribution and question prefixes of \ourname.}
    \label{fig:distribution-and-prefix}
\end{figure*}
\section{Details of Dataset Collection}
\label{sec:app_dataset}

To ensure broad visual diversity, we assembled a field team of more than 20 contributors distributed across four major cities spanning three continents. Each collector was tasked with capturing high-resolution photographs in supermarkets, cafés, museums, public transit hubs, and other everyday settings, following a shared shot-list that balanced lighting conditions, camera angles, and object categories. Distinct from prior datasets, all images were captured exclusively using three mainstream smart-glasses platforms—Ray-Ban Meta, Xiaomi Smart Glasses, and RayNeo AR glasses—rather than handheld phones or DSLRs. This decision provides both device heterogeneity (different optics, sensors, and ISP pipelines) and scenario fidelity, yielding data that is intrinsically aligned with real smart-glasses usage. The collection took place across varied times of day, weather conditions, and indoor/outdoor environments to further maximise visual diversity. Before entering the annotation workflow, all raw images were passed through a YOLO-based privacy filter\footnote{YOLO Model: \url{https://github.com/ultralytics/ultralytics}} to automatically redact faces, license plates, and other sensitive information that may inadvertently appear in in-the-wild capture. The curated images were then uploaded to a central server and annotated via a customised Label Studio\footnote{Label Studio: \url{https://labelstud.io}} interface with task-specific templates and scripted quality checks.

Every annotation underwent a dual-review process (collector → peer → project maintainer) to minimise label noise, standardise taxonomies, and enforce cross-annotator consistency. Category distributions were continuously monitored, and targeted sampling was employed to reduce long-tail imbalance. Collectively, these measures ensure that the resulting dataset is diverse, privacy-preserving, and faithfully representative of real-world smart-glasses VQA scenarios.

\section{Details of \ourname{}}
\label{sec:app_superglasses}

\subsection{Statistic Analysis}

\paragraph{Length distribution of questions and answers.} 
Figure~\ref{fig:length-distribution} shows the word-length distributions for all question and answer pairs in \benchname{}. Questions are typically concise and intent-driven, with a sharp peak around 7–10 words. In contrast, answers exhibit a much broader and flatter distribution, often extending beyond 50 words and occasionally reaching up to 100. This reflects the nature of smart-glasses interactions, where users issue short queries, but resolving them may require long-form reasoning, retrieval, or multi-hop tool usage. The distribution highlights the need for agents capable of handling both succinct visual prompts and compositional, evidence-grounded answers.

\paragraph{Distribution of question prefixes.}
Figure~\ref{fig:question-sunburst} visualizes the distribution of question openers in \benchname{} using a hierarchical sunburst chart. The vast majority of queries begin with \emph{“what”}, reflecting object-centric and identity-focused information needs common in smart-glasses usage. Other frequent prefixes include \emph{“how”}, \emph{“which”}, and \emph{“where”}, indicating procedural, choice-based, and location-oriented queries. Notably, the chart also reveals deeper branching structures, such as \emph{“what is”}, \emph{“how many”}, and \emph{“where is”}, capturing the compositional nature of natural questions. This distribution highlights the importance of handling a wide range of query intents—from factual lookups to spatial grounding and comparative reasoning.

\begin{figure*}[htbp]
    \centering
    \begin{subfigure}[b]{0.4\textwidth}
        \centering
        \includegraphics[width=\textwidth]{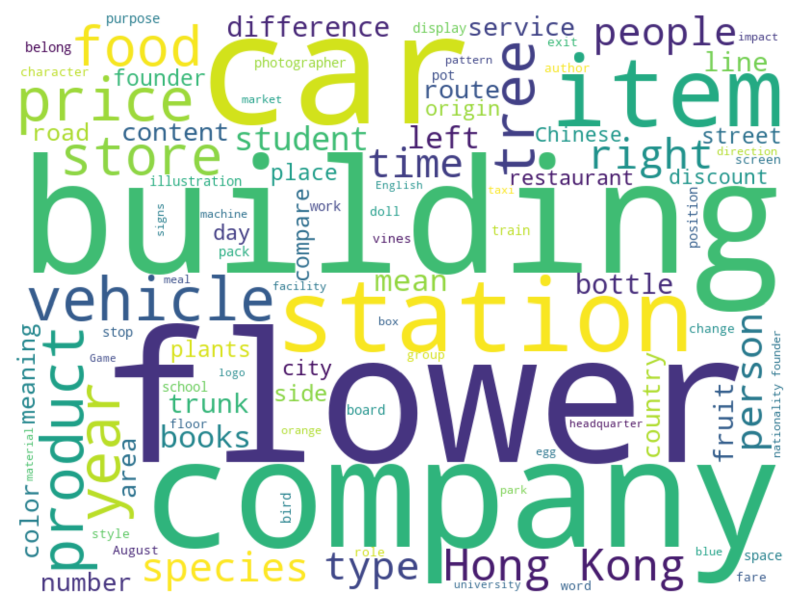}
        \caption{Common topics in questions.}
        \label{fig:word-cloud-question}
    \end{subfigure}
    \hfill
    \begin{subfigure}[b]{0.4\textwidth}
        \centering
        \includegraphics[width=\textwidth]{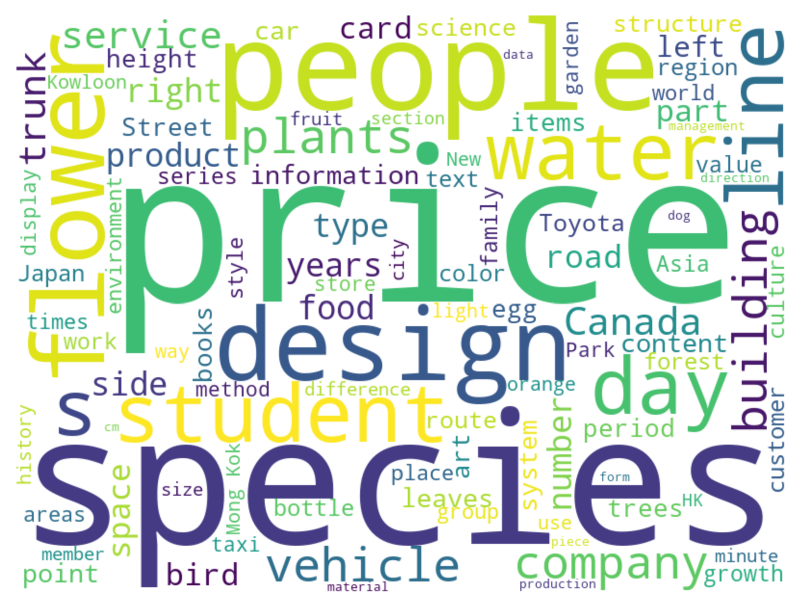}
        \caption{Common topics in answers.}
        \label{fig:word-cloud-answer}
    \end{subfigure}
    \vspace{5mm}
    \caption{The common topics in questions and answers of \ourname.
    }
    \label{fig:word-cloud}
\end{figure*}

\paragraph{Distribution of common topics in questions and answers.}
Figure~\ref{fig:word-cloud-question} and Figure~\ref{fig:word-cloud-answer} present word cloud visualizations of the most frequent content words appearing in questions and answers within \benchname{}. On the left, we observe that questions are heavily centered around visually grounded entities such as \emph{building}, \emph{flower}, \emph{car}, \emph{station}, and \emph{company}, reflecting the egocentric nature of smart-glasses interactions in daily environments. The right-side word cloud shows that answers often contain semantic labels like \emph{price}, \emph{species}, \emph{people}, \emph{design}, and \emph{water}, indicating a shift from visual recognition to external knowledge retrieval and classification. Together, the two distributions highlight the dataset's dual emphasis on visual object grounding and multi-hop, knowledge-intensive reasoning.

\subsection{Comparison with Other benchmarks} 
Early efforts such as WebQA \citep{chang2022webqa} and InfoSeek \citep{chen2023can} demonstrated the feasibility of multi-hop or information-seeking questions but were built on heterogeneous web crawls, lacked any notion of where the salient object sits in the frame, and provided no access to the underlying retrieval traces—making systematic analysis of failure cases or tool planning impossible. Subsequent benchmarks (MRAG-Bench \citep{hu2024mrag}, LIVEVQA \citep{fu2025livevqa}, MMSearch \citep{jiang2025mmsearch}, Dyn-VQA \citep{libenchmarking}, and CRAG-MM \citep{wang2025cragmm}) expanded linguistic coverage or introduced live-update scenarios, yet they still rely on static screenshots or curated news photos, treat the object’s spatial footprint as a latent variable, and offer only “black-box” evaluation of retrieval-augmented generation without exposing the decision path.

Our \benchname{} corpus addresses existing blind spots by capturing egocentric frames directly from smart-glasses users in real-world settings, precisely annotating \emph{object coverage} to enable spatially grounded reasoning, and recording the full \emph{search and tool-usage trajectory} to ensure that every answer is traceable to its supporting evidence. With comprehensive topical coverage across 14 distinct domains and 9 fine-grained query categories, this framework facilitates detailed result analysis, robust evaluation of retrieval and reasoning strategies, and realistic benchmarking of multimodal agents operating under the perceptual and interaction constraints of smart-glasses platforms.

\subsection{Examples in \textbf{\benchname{}}}

Figures~\ref{fig:query-metadata}, ~\ref{fig:hop1_hop2}, and ~\ref{fig:hop3_hop4} illustrate a four-hop example that begins with a shelf photograph of Campbell’s soup cans.  
The user asks \emph{“Which country is the renowned artist who painted this item from?”}.  
Answering requires chaining visual grounding with three successive text queries:

\begin{enumerate}[label=\textbf{Hop \arabic*:}]
  \item \textbf{Visual grounding.}  
        An image-search engine links the photo to the Wikipedia entry for \emph{Campbell’s}, confirming the product identity.

  \item \textbf{Artwork retrieval.}  
        A text search for the “famous painting depicting Campbell’s” returns the article on \emph{Campbell’s Soup Cans}, a 32-canvas pop-art series.

  \item \textbf{Artist identification.}  
        A follow-up query reveals that \emph{Campbell’s Soup Cans} was painted by pop-art pioneer \textbf{Andy Warhol}.

  \item \textbf{Nationality verification.}  
        Consulting Warhol’s biography shows he was an \textbf{American} artist, providing the target country.
\end{enumerate}

This case highlights three core properties of \ourname{}:  
(i) \emph{cross-modal grounding} (image $\rightarrow$ text),  
(ii) \emph{multi-document evidence aggregation} across four hops, and  
(iii) \emph{rich metadata} (device, location, difficulty, etc.) that enable fine-grained evaluation of multimodal reasoning systems.

\input{tables/appendix/domain_prompt}
\input{tables/appendix/direct_prompt}

\section{Details of \agentname{}}
\label{sec:app_superlens}
\subsection{Demand-Adaptive Answerer}
As shown in Figure 4 of the manuscript, we instruct the VLM to act as a domain router, identifying the domain of each question–image pair for domain-specific reasoning. 
The domain-recognition prompts are shown in Figure~\ref{fig:domain_prompt}, which outlines the scenarios for each predefined domain. 
To enable demand-adaptive retrieval, the model must also determine whether the question can be answered using its internal knowledge or requires external information. 
The prompt guiding this decision is illustrated in Figure~\ref{fig:direct_prompt}.
Here, we employ chain-of-thought prompting to activate the VLM’s reasoning capabilities fully.
Specifically, the VLM is instructed to identify precise object names in the image and explain their relationships to elements referenced in the question. 
If the VLM concludes that it has sufficient information to answer the question, it must present a comprehensive reasoning followed by the final answer.
Conversely, if key information is missing, it should halt reasoning and trigger retrieval, which is subsequently handled by the Dual-Lens Knowledge Retriever.

\subsection{Dual-Lens Knowledge Retriever}
Given an image-question pair $(v, q)$, the retriever aims to extract complementary external knowledge by decomposing the retrieval process into two parallel pathways: a visual lens and a textual lens.
First, a VLM-based Search Router is utilized to decide whether retrieval should rely on visual or textual signals, yielding two query sets:
\begin{equation}
    O, Q = \text{SearchRouter}(v, q),     
\end{equation}
where $O=\{o_1, ..., o_M\}$ are visual objects referenced in the input image and $Q = \{q_1, ..., q_N\}$ are textual queiries extracted from the orignal question.

For the visual branch, we employ an open-vocabulary Object Detector (Grounding DINO~\cite{liu2024grounding}) to localize the referred objects: $V = \{v_i = \text{ObjectDetector}(v, o_i) | o_i \in O\}$.
Each detected region $v_i$ is then used to index an image search engine, retrieving HTML pages or snippets associated with visually similar images: 
\begin{equation}
    H^{vis} = \bigcup_{i=1}^{M} H_i^{vis}, \quad H_i^{vis} = \text{ImageSearch}(v_i).
\end{equation}
Because textual queries may contain multi-hop reasoning, each query $q_i$ is further decomposed into sub-queries by a Query Decoupler, yielding an expanded set $Q^{\prime} = \{q_1^{\prime}, ..., q_{N^{\prime}}^{\prime}\}$.
After that, each sub-query is issued independently to a text search engine, resulting in retrieved webpages:
\begin{equation}
    H^{txt} = \bigcup_{j=1}^{N^{\prime}} H_j^{txt}, \quad H_j^{txt} = \text{TextSearch}(q_j^{\prime}).
\end{equation}
All retrieved webpages are merged into a unified set: $H = H^{vis} \cup H^{txt}$. In our setting, both image and text retrieval are conducted using a SerpApi-powered search engine\footnote{SerpApi: https://serpapi.com}, restricted to trustworthy web sources.

A Webpage Reader (ReaderLM-v2~\cite{wang2025readerlm}) is then used to parse raw HTML into clean, VLM-friendly text and segment it into manageable chunks: $C = \{c_1, c_2, ..., c_L\}$.
To align retrieved chunks with the input pair $(v, q)$, we compute a weighted multimodal relevance score for each chunk using a Multimodal Reranker~\cite{jina-reranker}. Since the reranker cannot jointly process both modalities at once, we evaluate relevance against the image and the question separately:
\begin{equation}
    S = \{s_l = w_1 \cdot \text{Reranker}(c_l, v) + w_2 \cdot \text{Reranker}(c_l, q) | c_l \in C\}
\end{equation}
where $s_l$ is the relevance score of chunk $c_l$ with input $(v, q)$; $w_1$ and $w_2$ are modality-balancing weights (set to 0.4 and 0.6, respectively).
Finally, we retain all chunks whose score exceeds a threshold $\tau_s$ (default 0.6) and select the top-$K$ chunks as the final RAG context.

\input{tables/appendix/domain-results}

\section{Details of Evaluation}
\label{sec:app_evaluation}

\subsection{LLM-as-Judge}
\label{sec:app_evaluator}
As shown in Figure~\ref{fig:evaluator_prompt}, we employ a structured evaluator prompt to assess answer correctness via an LLM-based evaluator, i.e., Qwen2.5-32B.
The evaluator is instructed to act as an expert QA judge and is given clear reasoning guidelines. 
It determines whether a predicted answer is accurate by comparing it against the ground truth, allowing for surface-level variation (e.g., paraphrasing) as long as the semantic content is preserved. 
The prompt enforces strict criteria: missing key details or including incorrect information results in a negative judgment. 
The output is a JSON object with a single Boolean field, \texttt{accuracy}, which enables consistent downstream aggregation and scoring.

\input{tables/appendix/evaluator_prompt}

\subsection{Settings of Direct Answering}
In the \textit{Direct Answering} setting, we adopt a concise VQA prompting scheme that forgoes chain-of-thought reasoning and external retrieval. 
The model is guided only by a lightweight system prompt that constrains its behavior (e.g., ``You are a helpful assistant that answers questions based on the provided image.'').
During inference, we directly concatenate the encoded image and user query with a brief user prompt (``Please give the answer within 1–2 sentences. Answer:''), which encourages the model to produce a short, image-grounded response relying solely on its vision–language understanding and inherent knowledge.
To reduce stochasticity in next-token prediction, we set the sampling temperature of all VLMs to 0.

\subsection{Settings of Heuristic RAG}
In the \textit{Heuristic RAG} setting, we employ a straightforward retrieval pipeline that relies solely on external search engines, without incorporating any learned retrieval components.
For each user query, we directly submit the textual query to Google Search and collect the top-ranked web snippets as candidate evidence.
In parallel, the input image is processed using Google Lens, which returns visually similar webpages along with their associated metadata.
All retrieved webpages are then passed through the Webpage Reader used in our \agentname{} system to extract clean textual content. 
The resulting snippets are concatenated into a lightweight context buffer and provided to the model without any reranking, filtering, or structured reasoning.
For \textit{Text RAG}, we use only the textual snippets retrieved from the original query.
For \textit{Image RAG}, we use only the image-relevant results returned by Google Lens.
Overall, this heuristic design approximates a naïve “search-then-answer” pipeline, allowing us to isolate how much performance can be attributed to simple external retrieval signals versus more advanced retrieval–augmented generation mechanisms.

\section{More Experimental Results}
\label{sec:app_experiment}

\input{tables/appendix/category-results}

In this section, we present the detailed scores of 26 leading VLMs alongside our \agentname{} across the top 10 image domains (Table~\ref{tab:doamin-results}) and 8 query categories (Table~\ref{tab:category-results}).
The results show that our method consistently achieves substantial improvements across all evaluated scenarios.
Notably, \agentname{}\ddag{} surpasses the performance of large-scale models such as Gemini 2.5 Pro and GPT-4o, both estimated to contain more than 400B parameters, on multiple domains and query categories.

\input{tables/appendix/dynamism-hops}
Table \ref{tab:dynamism-hops} presents detailed results across the Dynamism and Reasoning Hops dimensions. 
We observe that both \agentname{}\dag{} and \agentname{}\ddag{} achieve substantially larger improvements in the Fast-Changing setting than in the Static or Slow-Changing settings, highlighting the effectiveness of our approach for queries whose answers shift rapidly over time.
Additionally, while LLaMA-3.2-11B with direct answering performs poorly on 3-hop and 4-hop questions, its performance improves markedly when augmented with our method (\agentname{}\dag{}), showing gains of more than 17\% and 21\%, respectively. 
This pattern indicates that \agentname{} is highly capable of tackling complex, multi-step questions even when the backbone model is relatively weak, a capability largely attributable to the Query Decoupler and the fine-grained RAG pipeline.

\input{tables/appendix/llm_judge}
Furthermore, as table~\ref{tab:llm_judge} shows, we selected the LLM-as-judge metric for its consistency, scalability, and ability to capture semantic meaning—advantages that conventional measures such as lexical or embedding similarity often lack.
Meanwhile, we extend our metrics to include LLMs from diverse families, hybrid LLM-based evaluation, and human assessments to mitigate potential bias.
\input{tables/appendix/lang_distribution}
Also, We have incorporated the suggested multilingual analysis, evaluating performance on English (en), Chinese (zh), French (fr), and Japanese (jp) in our main experiments, as shown in Table~\ref{tab:lang_distribution}

\subsection{More Findings}
Humans and vision language models often diverge in their perception of question difficulty and the necessity of external retrieval. 
Tasks that appear easy for humans may challenge models without additional context, while some seemingly complex questions for humans are handled well by models through memorized patterns. 
Notably, stronger models like GPT-4o and Gemini 2.5 Pro rely less on retrieval, maintaining high accuracy even without external input, suggesting greater internal world knowledge. 
In contrast, weaker models benefit more from retrieval but are also more sensitive to irrelevant results. 

\subsection{Case Study}
To better understand how our system behaves across different multimodal reasoning scenarios, we conduct a targeted case study that examines both its successes and failures in real examples. 
Specifically, we present two success cases and two failure cases to analyze the end-to-end decision process, from tool invocation to query decoupling to evidence aggregation. 
The success cases highlight situations where the model accurately identifies the appropriate retrieval modality (image search vs. text search) and constructs well-formed queries that lead to reliable answers. 
In contrast, the failure cases reveal two major failure modes: incorrect selection of the search tool and poorly structured query decoupling. 
Together, these examples provide a fine-grained, qualitative view of how the system makes decisions, what it does well, and where the current limitations lie.

\subsubsection{Success Case}
\begin{figure*}
    \centering
    \includegraphics[width=\linewidth]{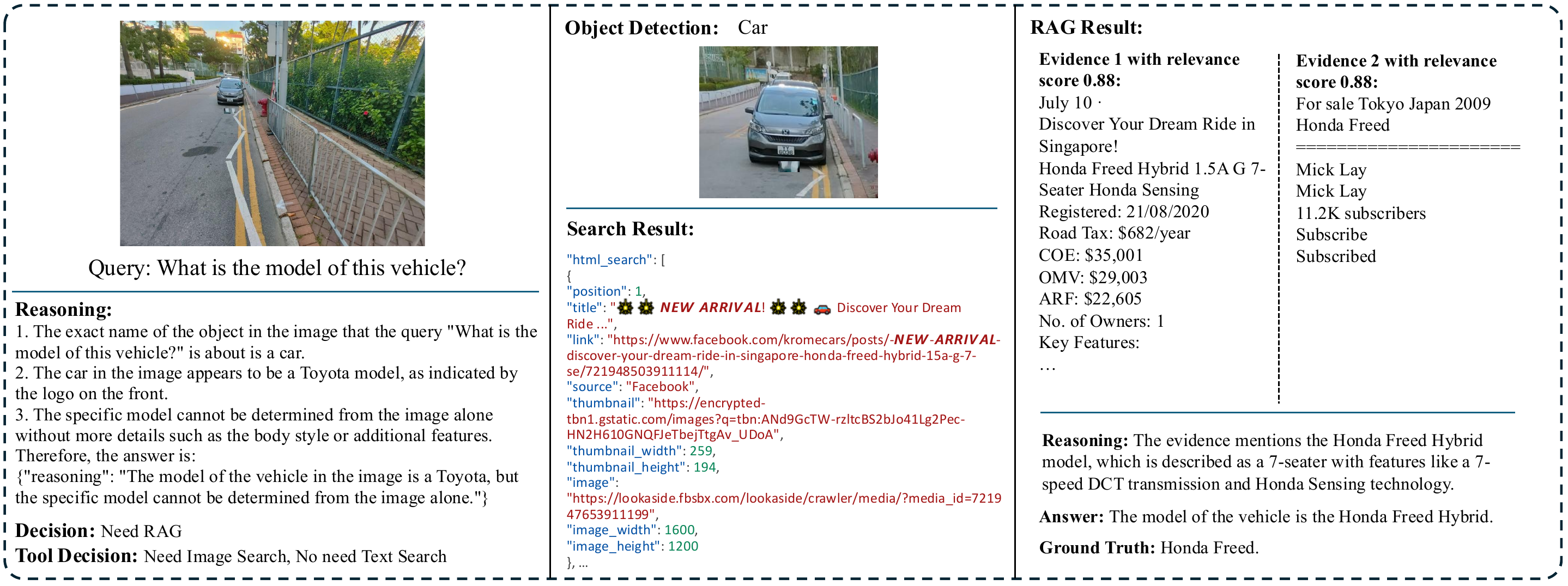}
    \caption{Success case of using image search.}
    \label{fig:corr_case_1}
\end{figure*}

\begin{figure*}
    \centering
    \includegraphics[width=\linewidth]{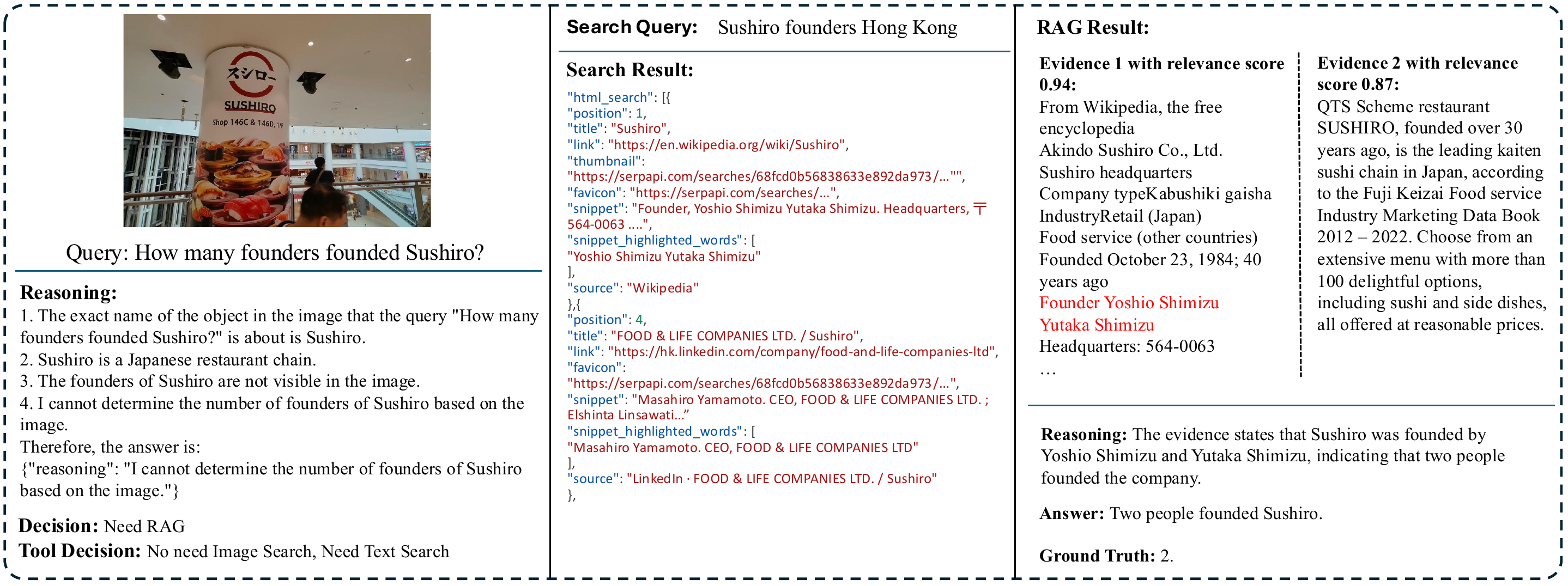}
    \caption{Success case of using text search.}
    \label{fig:corr_case_2}
\end{figure*}

To illustrate how our system behaves under ideal conditions, we present two representative success cases, each showcasing a different but appropriate tool invocation strategy. 
In the first case (Figure~\ref{fig:corr_case_1}), the model is asked to identify the vehicle model from an image. 
The system first performs precise visual grounding by extracting the relevant region through object detection, and then issues an image-based search query tailored to the cropped vehicle. 
This example illustrates that the model not only selects the correct modality (image search instead of text search) but also constructs a semantically faithful and discriminative query, enabling the retriever to return the correct result for the “Honda Freed Hybrid”.

In contrast, the second case (Figure~\ref{fig:corr_case_2}) demonstrates a query whose answer is not visually observable, i.e., the number of founders of the Sushiro restaurant chain. 
Here, the model successfully recognizes that the image provides insufficient information and therefore switches to text search. 
It generates a concise and meaningful search query (“Sushiro founders Hong Kong”) aligned with the question intent, retrieves high-quality textual evidence, and produces the correct answer (“two founders”). 
Together, these two examples show that when the system correctly selects the appropriate search tool, forms a well-structured query, and extracts modality-specific evidence, it is capable of producing robust and accurate answers across heterogeneous question types.

\subsubsection{Failure Case}

\begin{figure*}
    \centering
    \includegraphics[width=\linewidth]{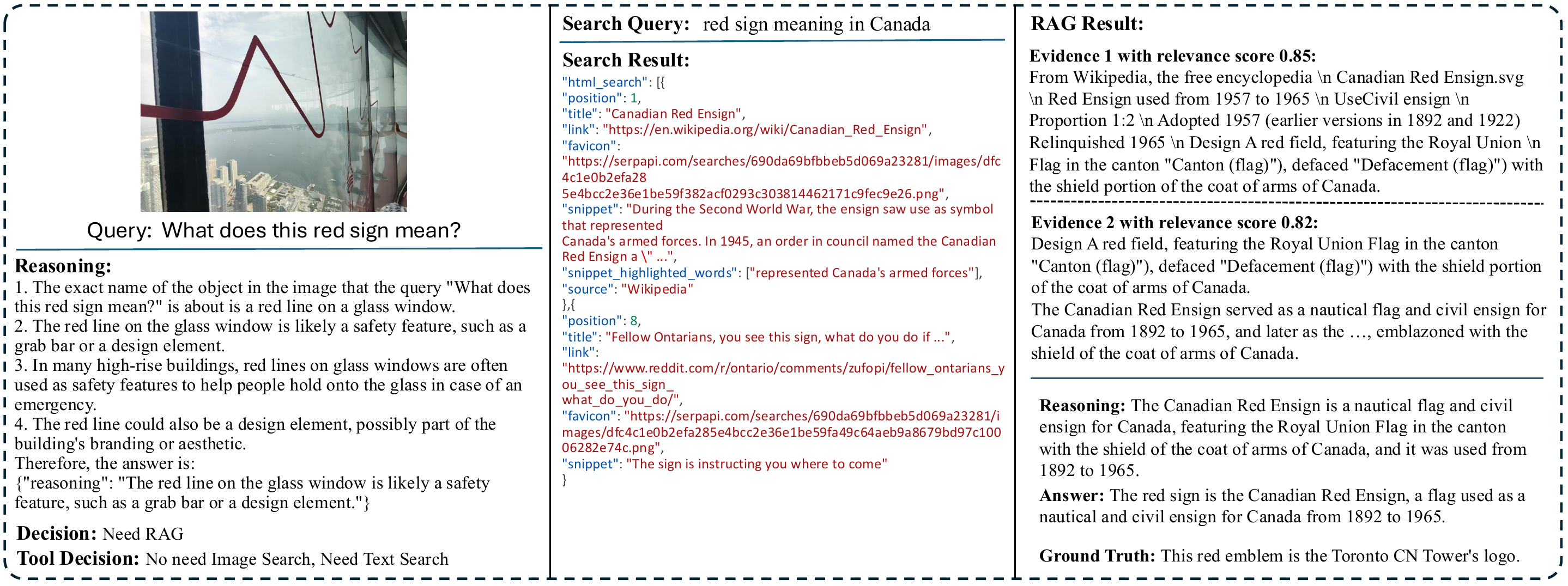}
    \caption{Failure case caused by incorrect search tool selection.}
    \label{fig:err_case_1}
\end{figure*}

\begin{figure*}
    \centering
    \includegraphics[width=\linewidth]{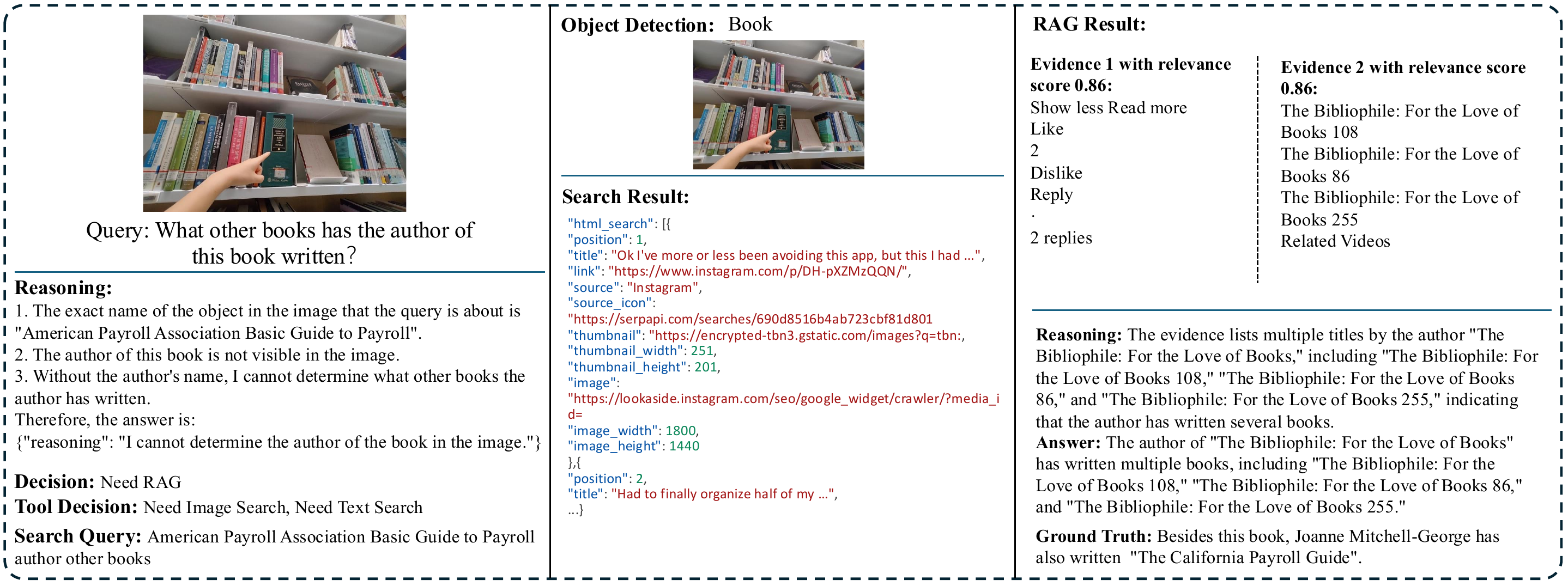}
    \caption{Failure case caused by incorrect search query generation.}
    \label{fig:err_case_2}
\end{figure*}

In addition to the success examples, we further present two failure cases (Figures~\ref{fig:err_case_1} and~\ref{fig:err_case_2}) to highlight the system’s failure modes and the underlying causes.

The first case (Figure~\ref{fig:err_case_1}) illustrates a mis-selection of the search tool. 
The question asks about a red symbol visible on a glass window. 
Instead of performing image-based search, which is necessary because the symbol is visually specific, hard to describe verbally, and not uniquely identifiable by text, the system mistakenly triggers text search with the query “red sign meaning in Canada.” 
This textual query is overly broad and semantically disconnected from the object in the image, leading the retriever to return irrelevant knowledge about the “Canadian Red Ensign” flag. 
This failure demonstrates that when the model misjudges the modality of the problem and chooses the wrong retrieval channel, even a formally well-constructed text query cannot compensate for the mismatch between modality and information need.

The second case (Figure~\ref{fig:err_case_2}) showcases a different type of failure: incorrect search query generation. 
The question asks, “What other books has the author of this book written?” The image clearly shows only the book cover and bookshelf, meaning the system should first infer the author’s name visually and then construct a clean, author-centric search query. However, the model generates a noisy and poorly structured query (“American Payroll Association Basic Guide to Payroll author other books”), which conflates the book title with the intent of the question and fails to isolate the key entity, i.e., the author. As a result, the retriever returns incomplete or misleading evidence. 
This example indicates that even when the correct tool (hybrid image and text search) is selected, an ill-formed query severely degrades retrieval quality.

Together, these two failure cases reveal complementary weaknesses of the system: the first arises from incorrect tool selection, while the second reflects deficient query formulation. 
These failures emphasize the importance of modality-aware decision-making and precise query construction for building robust multimodal RAG systems.

\input{tables/appendix/datacase}

%% file: sections/1_RelatedWork.tex
\section{Related Works}
\label{sec:related_work}

\subsection{Retrieval-Augmented VQA}
Retrieval-augmented Visual Question Answering (RA-VQA)~\citep{lin2023fine} extends knowledge-based VQA by coupling visual grounding with external retrieval from text corpora, KGs, or the open web (e.g., OK-VQA~\citep{wang2023filling}, A-OKVQA~\citep{schwenk2022okvqa}, WebQA~\citep{chang2022webqa}). Methods span knowledge-augmented transformers and RAG-enhanced LMMs to agentic/tool-use frameworks (ViperGPT~\citep{suris2023vipergpt}, CRAG-MM~\citep{wang2025cragmm}) that decide when to search, OCR, or compose multi-step programs. Recent systems adopt multi-turn, multi-hop policies that coordinate text–image retrieval and unify retrieval–generation end-to-end, learning not only to answer but to find, select, and justify evidence. Distinct from smart glasses settings, targets often occupy a very small portion of egocentric frames, making grounding and retrieval more challenging; moreover, there remains a notable lack of RA-VQA datasets specifically designed for smart glasses egocentric data.

\subsection{Vision Language Model-based Agents}
With the rapid advancement of large language models (LLMs)~\citep{wu2025mmsearch, jiang2025hibench}, multimodal agents have become increasingly capable of perceiving, reasoning, and acting across diverse modalities. 
Vision Language model (VLM)–based agents~\citep{wang2025mllm} have progressed from tool-orchestration wrappers to systems MLLM-based agents have evolved from tool orchestrators (e.g., Visual ChatGPT~\citep{wu2023visual}, HuggingGPT~\citep{shen2023hugginggpt}) to systems that perceive–plan–act with retrieval, using prompting policies and program synthesis (ViperGPT~\citep{suris2023vipergpt}) to decompose goals and call OCR/search/vision tools. Moving beyond static images, device/UI agents operate real apps from pixels (AppAgent~\citep{zhang2025appagent}, SeeClick~\citep{cheng2024seeclick}), while embodied models (PaLM-E~\citep{driess2023palm}, RT-2~\citep{zitkovich2023rt}) couple perception with action. For knowledge-heavy tasks, QA-Dragon~\citep{jiang2025qadragon} introduces a query-aware dynamic RAG, routing across text/image retrieval for multi-turn, multi-hop evidence. However, a clear gap remains: dedicated agentic stacks, purpose-built for smart glasses' egocentric inputs, are still lacking.

%% file: tables/appendix/domain_prompt.tex
\begin{figure*}[h]
\begin{mybox}{Domain Router Prompt}
\textbf{System Prompt:}

You are a visual assistant that identifies the question domain based on the query and image.
The question domain should be one of the following: [``Food'', ``Shopping'', ``Plant'', ``Public Service'', ``Culture'', ..., ``Other''].\\

- Food: Questions about dishes, ingredients, nutrition, cooking methods, or the cultural/industrial origin of food items.\\
- Shopping: Questions about consumer goods or published media—price, specifications, packaging, availability, editions, or author/publisher details.\\
...\\
If you are not sure about the question domain, you should return ``Other''.\\

\textbf{User Prompt:}\\
Given the \verb|<image>| and query text: \{query\}.\\
Output your predicted domain in JSON format, like
\{``domain'': \verb|<domain>|\}
\end{mybox}
\caption{The prompt used for domain recognition.}
\label{fig:domain_prompt}
\end{figure*}

%% file: tables/appendix/direct_prompt.tex
\begin{figure*}[h]
\begin{mybox}{Direct Answer Generation Prompt}
\textbf{System Prompt:}

You are a visual assistant tasked with addressing the user's query for the image based on your inherent knowledge.\\
General Reasoning Guidelines:\\
1. Generate step-by-step reasoning to address the query using evidence from the image and your knowledge...
Stop reasoning once you have enough information to answer, or you find that necessary information is lacking.\\
2. In your reasoning, identify the exact object that the query is about by its exact name...\\
3. If the query involves multiple objects or relationships, dedicate one reasoning step to each object or relationship, and then summarize the result in a final step.\\
4. If you find that necessary information is lacking, explicitly state: ``I have no knowledge about \verb|<lacking_knowledge>|''\\

Domain Reasoning Guidelines:\\
...\\

\textbf{User Prompt:}\\
Given the \verb|<image>|, please conduct step-by-step reasoning to address the query: \{query\}\\
Image metadata: The location of the image is \{location\}.

Output Format:\\
1. The exact name of the object in the image that the query is about is \verb|<specific_object_name>|.\\
2. Then, ...\\
3. Therefore, the answer is ...\\
Output Summary in JSON format:
\{``reasoning'': \verb|<summary_reasoning_string>|, ``answer'': \verb|<answer>|\}
\end{mybox}
\caption{The prompt used for direct answer generation.}
\label{fig:direct_prompt}
\end{figure*}

%% file: tables/appendix/domain-results.tex
\begin{table*}[t]
\caption{Detailed scores on top-10 domains of \ourname{}.}
\label{tab:doamin-results}
\resizebox{\textwidth}{!}{
\begin{tabular}{lcccccccccc}
\toprule
\textbf{Model}             
& \textbf{Plant}     
& \textbf{\makecell[c]{Public \\ Service}}    
& \textbf{Food}     
& \textbf{Shopping}  
& \textbf{Translation}  
& \textbf{Transport} 
& \textbf{Culture}  
& \textbf{Navigation} 
& \textbf{Animal}     
& \textbf{Education}                     \\
\midrule
\rowcolor{gray!20} \multicolumn{11}{c}{\textit{{\raisebox{-0.2\height}{\includegraphics[height=1.1em]{figures/logo/meta.pdf}}}~~\textbf{Meta-Rayben Smart Glasses}}} \\
\midrule
LLaMA-3.2-11B              & 22.49 & 24.58 & 18.82 & 28.27 & 25.11 & 17.24 & 25.13 & 30.46 & 18.35 & 30.00 \\
LLaMA-3.2-90B              & 34.32 & 33.00 & 26.20 & 34.60 & 34.04 & 26.11 & 33.51 & 35.06 & 25.32 & 37.50 \\

\midrule
\rowcolor{gray!15} \multicolumn{11}{c}{{\raisebox{-0.2\height}{\includegraphics[height=1.3em]{figures/logo/rayneo.pdf}}}~\textit{\textbf{RayNeo Smart Glasses}}} \\
\midrule
Qwen2.5-VL-3B                & 26.04 & 28.96 & 20.30 & 29.96 & 25.11 & 20.20 & 25.13 & 33.91 & 10.76 & 40.00 \\
Qwen2.5-VL-7B                & 31.07 & 34.01 & 22.51 & 35.44 & 39.57 & 25.12 & 26.70 & 48.28 & 25.95 & 45.00 \\
Qwen2.5-VL-32B               & 37.87 & 36.36 & 29.89 & 40.08 & 43.83 & 26.11 & 36.65 & 44.83 & 28.48 & 50.00 \\
Qwen2.5-VL-72B               & 37.87 & 39.73 & 28.78 & 42.62 & 43.40 & 31.03 & 34.03 & 48.85 & 31.65 & 46.25 \\
\midrule
\rowcolor{gray!15} \multicolumn{11}{c}{{\raisebox{-0.2\height}{\includegraphics[height=1.1em]{figures/logo/xiaomi.pdf}}}~~\textit{\textbf{XiaoMi Smart Glasses}}} \\
\midrule
MiMo-VL-7B                   & 26.63 & 25.93 & 20.30 & 28.69 & 21.70 & 15.76 & 27.23 & 32.76 & 18.35 & 37.50 \\

\midrule

\rowcolor{gray!15} \multicolumn{11}{c}{{\raisebox{-0.4\height}{\includegraphics[height=1.1em]{figures/logo/open-source.pdf}}}~~\textit{\textbf{Open-sourced VLMs}}} \\
\midrule
Phi-3-Vision-4B              & 18.34 & 19.53 & 15.13 & 18.57 & 16.60 & 11.33 & 16.23 & 21.84 & 11.39 & 27.50 \\
InternVL3-8B                 & 24.26 & 30.98 & 19.19 & 32.49 & 28.51 & 15.27 & 26.18 & 36.78 & 22.15 & 40.00 \\
GLM-4.1V-9B                  & 22.78 & 23.57 & 21.03 & 27.43 & 22.98 & 19.70 & 27.23 & 27.01 & 18.35 & 35.00 \\
LLaVA-v1.5-7B                & 11.54 & 9.76  & 11.44 & 12.66 & 7.23  & 6.90  & 12.57 & 13.22 & 6.96  & 12.50 \\
LLaVA-v1.5 -13B              & 13.61 & 10.44 & 12.55 & 16.03 & 5.11  & 9.36  & 14.66 & 12.07 & 9.49  & 15.00 \\
LLaVA-Onevision-0.5B         & 10.65 & 16.50 & 12.92 & 18.14 & 7.66  & 12.81 & 14.14 & 16.67 & 13.29 & 25.00 \\
LLaVA-Onevision-7B           & 19.82 & 21.21 & 15.87 & 24.47 & 17.45 & 16.26 & 20.42 & 31.61 & 14.56 & 36.25 \\
DeepSeek-VL2-3B              & 17.46 & 20.54 & 17.71 & 21.94 & 16.60 & 18.72 & 19.37 & 29.31 & 15.19 & 35.00 \\
DeepSeek-VL2-16B             & 24.26 & 23.23 & 21.03 & 28.27 & 21.70 & 18.72 & 24.08 & 32.76 & 17.09 & 33.75 \\
DeepSeek-VL2-27B             & 23.08 & 28.96 & 21.77 & 31.65 & 25.53 & 17.73 & 26.18 & 33.91 & 18.99 & 40.00 \\
\midrule
\rowcolor{gray!15} \multicolumn{11}{c}{{\raisebox{-0.2\height}{\includegraphics[height=1.1em]{figures/logo/chat-gpt.pdf}}}~~\textit{\textbf{Proprietary VLMs}}} \\
\midrule
GPT-4o                       & \textbf{46.75} & 43.10 & 40.22 & 44.30 & 42.55 & 33.99 & 38.74 & 44.83 & 39.87 & 53.75 \\
Claude 4 Sonnet              & 31.95 & 37.37 & 38.38 & 42.19 & 41.28 & 29.56 & 36.13 & 43.10 & 27.22 & 53.75 \\
Gemini 2.5 Pro               & 40.83 & 46.13 & \textbf{42.80} & 46.41 & 38.30 & \textbf{41.38} & \textbf{39.79} & 48.85 & \textbf{36.08} & 53.75 \\
\midrule
\rowcolor{gray!15} \multicolumn{11}{c}{{\raisebox{-0.2\height}{\includegraphics[height=1.1em]{figures/logo/search.pdf}}}~~\textit{\textbf{Heuristic RAG}}} \\
\midrule
LLaMA-3.2-11B                & 14.50 & 15.49 & 14.02 & 24.89 & 15.32 & 14.29 & 19.90 & 20.69 & 10.76 & 36.25 \\
LLaMA-3.2-11B                & 11.54 & 12.12 & 12.92 & 14.77 & 12.77 & 12.32 & 16.23 & 14.94 & 8.23  & 20.00 \\
LLaMA-3.2-11B                & 12.72 & 10.77 & 14.39 & 16.46 & 14.04 & 13.30 & 19.90 & 19.54 & 9.49  & 26.25 \\
Qwen2.5-VL-7B                & 15.09 & 19.87 & 19.19 & 21.52 & 20.00 & 13.30 & 20.94 & 26.44 & 12.66 & 41.25 \\
Qwen2.5-VL-7B                & 16.57 & 15.15 & 15.50 & 17.72 & 17.87 & 13.30 & 17.80 & 19.54 & 10.13 & 26.25 \\
Qwen2.5-VL-7B                & 16.57 & 17.17 & 14.39 & 21.10 & 17.87 & 12.32 & 19.37 & 20.69 & 12.03 & 26.25 \\
\midrule
\textbf{\agentname{}\dag}~(Ours) 
& \gaincell{33.43}{10.94}
& \gaincell{36.03}{11.45}
& \gaincell{32.10}{13.28}
& \gaincell{48.52}{20.25}
& \gaincell{33.62}{8.51}
& \gaincell{33.00}{15.76}
& \gaincell{40.31}{15.18}
& \gaincell{48.28}{17.82}
& \gaincell{36.08}{17.73}
& \gaincell{41.25}{11.25}
\\
\midrule
\textbf{\agentname{}\ddag}~(Ours) 
& \gaincell{39.64}{8.57}
& \gaincell{\textbf{47.47}}{13.46}
& \gaincell{34.32}{11.81}
& \gaincell{\textbf{47.68}}{12.24}
& \gaincell{\textbf{51.91}}{12.34}
& \gaincell{38.42}{13.30}
& \gaincell{38.74}{12.04}
& \gaincell{\textbf{56.90}}{8.62}
& \gaincell{34.81}{8.86}
& \gaincell{\textbf{56.25}}{11.25}
\\
\bottomrule
\end{tabular}
}
\end{table*}

%% file: tables/appendix/evaluator_prompt.tex
\begin{figure*}[h]
\begin{mybox}{Evaluator Prompt}
\textbf{System Prompt:}

You are an expert evaluator of question-answering systems. 

\textbf{User Prompt:}\\
\textit{General Reasoning Guidelines:}
``Your task is to determine if a prediction correctly answers a question based on the ground truth.''\\
\textit{Rules:}
    \begin{enumerate}
    \item The prediction is correct if it captures all the key information from the ground truth.
    \item The prediction is correct even if phrased differently as long as the meaning is the same.
    \item The prediction is incorrect if it contains incorrect information or is missing essential details.
    ``Output a JSON object with a single field `accuracy' whose value is true or false.''
    \end{enumerate}
    \textit{Question:} \{query\}, \textit{Ground Truth:} \{answer\}, \textit{Prediction:} \{prediction\}
\end{mybox}
\caption{The prompt used for answer evaluation.}
\label{fig:evaluator_prompt}
\end{figure*}

%% file: tables/appendix/category-results.tex
\begin{table*}[t]
\caption{Detailed scores on query categories of \ourname{}.}
\resizebox{\textwidth}{!}{
\begin{tabular}{lcccccccc}
\toprule
\textbf{Model}             
& \textbf{Aggregation}     
& \textbf{Comparison}    
& \textbf{\makecell[c]{Factual \\ Knowledge}}     
& \textbf{Multi-hop}  
& \textbf{Reasoning}  
& \textbf{\makecell[c]{Simple \\ Recognition}}
& \textbf{\makecell[c]{Spatial \\ Reasoning}}
& \textbf{\makecell[c]{Temporal \\ Understanding}}             
\\
\midrule
\rowcolor{gray!20} \multicolumn{9}{c}{\textit{{\raisebox{-0.2\height}{\includegraphics[height=1.1em]{figures/logo/meta.pdf}}}~~\textbf{Meta-Rayben Smart Glasses}}} \\
\midrule
LLaMA-3.2-11B              &  19.42 & 21.23 & 18.09 & 16.79 & 32.38 & 24.53 & 33.94 & 21.54 \\
LLaMA-3.2-90B              &  26.62 & 32.88 & 26.50 & 23.80 & 38.95 & 34.08 & 38.53 & 32.31 \\

\midrule
\rowcolor{gray!15} \multicolumn{9}{c}{{\raisebox{-0.2\height}{\includegraphics[height=1.3em]{figures/logo/rayneo.pdf}}}~\textit{\textbf{RayNeo Smart Glasses}}} \\
\midrule
Qwen2.5-VL-3B                &  22.30 & 28.08 & 17.52 & 17.52 & 31.03 & 30.43 & 31.19 & 24.62 \\
Qwen2.5-VL-7B                &  30.22 & 34.93 & 21.94 & 23.94 & 41.48 & 39.33 & 34.86 & 30.77 \\
Qwen2.5-VL-32B               &  29.50 & 39.73 & 26.21 & 27.30 & 48.57 & 42.13 & 39.45 & 38.46 \\
Qwen2.5-VL-72B               &  37.41 & 42.47 & 28.21 & 28.76 & 50.25 & 42.32 & 37.61 & 35.38 \\
\midrule
\rowcolor{gray!15} \multicolumn{9}{c}{{\raisebox{-0.2\height}{\includegraphics[height=1.1em]{figures/logo/xiaomi.pdf}}}~~\textit{\textbf{XiaoMi Smart Glasses}}} \\
\midrule
MiMo-VL-7B                   &  25.18 & 27.40 & 16.67 & 15.77 & 29.51 & 29.12 & 35.78 & 27.69 \\

\midrule

\rowcolor{gray!15} \multicolumn{9}{c}{{\raisebox{-0.4\height}{\includegraphics[height=1.1em]{figures/logo/open-source.pdf}}}~~\textit{\textbf{Open-sourced VLMs}}} \\
\midrule
Phi-3-Vision-4B              &  12.23 & 27.40 & 11.68 & 11.09 & 24.79 & 18.63 & 24.77 & 21.54 \\
InternVL3-8B                 &  24.46 & 32.19 & 16.81 & 19.85 & 36.76 & 30.99 & 34.86 & 30.77 \\
GLM-4.1V-9B                  &  22.30 & 28.77 & 16.52 & 16.35 & 31.87 & 27.72 & 24.77 & 21.54 \\
LLaVA-v1.5-7B                &  10.79 & 17.12 & 6.41  & 7.88  & 13.15 & 11.89 & 20.18 & 12.31 \\
LLaVA-v1.5 -13B              &  13.67 & 20.55 & 7.98  & 8.03  & 13.49 & 12.17 & 24.77 & 20.00 \\
LLaVA-Onevision-0.5B         &  13.67 & 17.81 & 7.98  & 8.18  & 16.69 & 16.57 & 18.35 & 18.46 \\
LLaVA-Onevision-7B           &  20.86 & 30.82 & 13.53 & 14.45 & 26.14 & 23.69 & 34.86 & 30.77 \\
DeepSeek-VL2-3B              &  17.27 & 28.08 & 11.54 & 11.39 & 26.64 & 23.69 & 26.61 & 24.62 \\
DeepSeek-VL2-16B             &  23.74 & 30.14 & 14.67 & 15.62 & 29.51 & 29.96 & 27.52 & 24.62 \\
DeepSeek-VL2-27B             &  29.50 & 34.93 & 22.22 & 19.85 & 34.57 & 32.96 & 37.61 & 32.31 \\
\midrule
\rowcolor{gray!15} \multicolumn{9}{c}{{\raisebox{-0.2\height}{\includegraphics[height=1.1em]{figures/logo/chat-gpt.pdf}}}~~\textit{\textbf{Proprietary VLMs}}} \\
\midrule
GPT-4o                       &  34.53 & 41.78 & 37.89 & 34.89 & 50.08 & 44.48 & \textbf{52.29} & 38.46 \\
Claude 4 Sonnet              &  35.25 & 38.36 & 29.06 & 29.49 & 49.75 & 41.01 & 41.28 & 43.08 \\
Gemini 2.5 Pro               &  \textbf{41.73} & \textbf{51.37} & \textbf{38.18} & \textbf{38.39} & 50.25 & 45.13 & 43.12 & 36.92 \\
\midrule
\rowcolor{gray!15} \multicolumn{9}{c}{{\raisebox{-0.2\height}{\includegraphics[height=1.1em]{figures/logo/search.pdf}}}~~\textit{\textbf{Heuristic RAG}}} \\
\midrule
LLaMA-3.2-11B                &  11.51 & 17.81 & 12.68 & 11.97 & 23.27 & 19.76 & 22.02 & 26.15 \\
LLaMA-3.2-11B                &  12.95 & 21.92 & 12.25 & 9.49  & 16.53 & 13.01 & 14.68 & 21.54 \\
LLaMA-3.2-11B                &  15.11 & 25.34 & 12.68 & 9.64  & 18.21 & 14.61 & 14.68 & 26.15 \\
Qwen2.5-VL-7B                &  16.55 & 20.55 & 13.82 & 12.12 & 27.15 & 22.94 & 16.51 & 24.62 \\
Qwen2.5-VL-7B                &  12.23 & 19.18 & 13.82 & 10.07 & 20.57 & 17.04 & 15.60 & 23.08 \\
Qwen2.5-VL-7B                &  12.23 & 20.55 & 13.68 & 9.78  & 20.57 & 18.63 & 19.27 & 23.08 \\
\midrule
\textbf{\agentname{}\dag}~(Ours) 
& \gaincell{27.34}{7.92}
& \gaincell{43.84}{22.61}
& \gaincell{37.61}{19.52}
& \gaincell{30.80}{14.01}
& \gaincell{42.16}{9.78}
& \gaincell{36.99}{12.46}
& \gaincell{43.12}{9.18}
& \gaincell{32.31}{10.77}
\\
\midrule
\textbf{\agentname{}\ddag}~(Ours) 
& \gaincell{39.57}{9.35}
& \gaincell{45.89}{10.96}
& \gaincell{36.47}{14.53}
& \gaincell{34.74}{10.80}
& \gaincell{\textbf{52.28}}{10.80}
& \gaincell{\textbf{49.34}}{10.01}
& \gaincell{44.95}{10.09}
& \gaincell{\textbf{44.62}}{13.85}
\\
\bottomrule
\end{tabular}
}
\label{tab:category-results}
\end{table*}

%% file: tables/appendix/dynamism-hops.tex
\begin{table*}[t]
\centering
\caption{Detailed scores on dynamism and hop dimensions of SuperGlasses.}
\resizebox{0.95\textwidth}{!}{
\begin{tabular}{lcccccccc}
\toprule
\multirow{2}{*}{\textbf{Model}} & \multirow{2}{*}{\textbf{Serach Type}} & \multicolumn{3}{c}{\textbf{Dynamism}} & \multicolumn{4}{c}{\textbf{Reasoning Hops}} \\
\cmidrule{3-9}
                         &    & Static          & Slow-Changing         & Fast-Changing     & 1-hop     & 2-hop    & 3-hop     & 4-hop  \\
\midrule
\rowcolor{gray!20} \multicolumn{9}{c}{\textit{{\raisebox{-0.2\height}{\includegraphics[height=1.1em]{figures/logo/meta.pdf}}}~~\textbf{Meta-Rayben Smart Glasses}}} \\
\midrule
LLaMA-3.2-11B       & Direct Answer          & 25.21 & 13.66 & 13.92 & 26.35 & 23.53 & 16.90 & 4.35  \\
LLaMA-3.2-90B      & Direct Answer           & 32.80 & 22.36 & 26.29 & 34.55 & 29.41 & 24.80 & 17.39 \\

\midrule
\rowcolor{gray!15} \multicolumn{9}{c}{{\raisebox{-0.2\height}{\includegraphics[height=1.3em]{figures/logo/rayneo.pdf}}}~\textit{\textbf{RayNeo Smart Glasses}}} \\
\midrule
Qwen2.5-VL-3B       & Direct Answer          & 42.59 & 25.77 & 18.81 & 29.02 & 21.85 & 17.69 & 13.04 \\
Qwen2.5-VL-7B      &  Direct Answer          & 34.64 & 23.60 & 21.13 & 36.92 & 27.73 & 23.70 & 17.39 \\
Qwen2.5-VL-32B     &  Direct Answer          & 37.93 & 26.09 & 30.41 & 40.50 & 33.61 & 27.33 & 21.74 \\
Qwen2.5-VL-72B    & Direct Answer            & 38.85 & 27.95 & 32.99 & 41.29 & 36.13 & 29.07 & 21.74 \\
\midrule
\rowcolor{gray!15} \multicolumn{9}{c}{{\raisebox{-0.2\height}{\includegraphics[height=1.1em]{figures/logo/xiaomi.pdf}}}~~\textit{\textbf{XiaoMi Smart Glasses}}} \\
\midrule
MiMo-VL-7B        & Direct Answer            & 26.08 & 16.77 & 20.62 & 28.96 & 15.97 & 16.90 & 13.04 \\

\midrule

\rowcolor{gray!15} \multicolumn{9}{c}{{\raisebox{-0.4\height}{\includegraphics[height=1.1em]{figures/logo/open-source.pdf}}}~~\textit{\textbf{Open-sourced VLMs}}} \\
\midrule
Phi-3-Vision-4B   & Direct Answer            & 18.05 & 12.42 & 13.40 & 19.98 & 16.81 & 10.58 & 13.04 \\
InternVL3-8B       & Direct Answer           & 28.25 & 17.39 & 20.10 & 29.99 & 25.21 & 19.27 & 21.74 \\
GLM-4.1V-9B      & Direct Answer             & 25.01 & 13.66 & 19.59 & 26.53 & 22.69 & 17.38 & 13.04 \\
LLaVA-v1.5-7B    & Direct Answer             & 10.93 & 8.07  & 7.22  & 11.35 & 12.61 & 7.74  & 8.70  \\
LLaVA-v1.5 -13B   & Direct Answer            & 12.63 & 4.35  & 8.25  & 13.36 & 13.45 & 7.27  & 8.70  \\
LLaVA-Onevision-0.5B   & Direct Answer       & 14.32 & 12.42 & 8.25  & 15.79 & 14.29 & 8.37  & 8.70  \\
LLaVA-Onevision-7B    & Direct Answer        & 21.82 & 14.91 & 17.01 & 23.92 & 21.01 & 13.59 & 13.04 \\
DeepSeek-VL2-3B    & Direct Answer           & 20.37 & 16.15 & 15.98 & 23.13 & 15.13 & 11.85 & 17.39 \\
DeepSeek-VL2-16B   & Direct Answer           & 24.67 & 21.12 & 18.04 & 27.50 & 19.33 & 15.96 & 8.70  \\
DeepSeek-VL2-27B   & Direct Answer           & 26.42 & 22.36 & 20.10 & 28.48 & 23.53 & 19.12 & 13.04 \\
\midrule
\rowcolor{gray!15} \multicolumn{9}{c}{{\raisebox{-0.2\height}{\includegraphics[height=1.1em]{figures/logo/chat-gpt.pdf}}}~~\textit{\textbf{Proprietary VLMs}}} \\
\midrule
GPT-4o          & Direct Answer              & 43.88 & 29.19 & 31.44 & 44.87 & 42.86 & 34.44 & 30.43 \\
Claude 4 Sonnet    & Direct Answer           & 37.93 & 29.19 & 35.57 & 40.32 & 37.82 & 29.07 & 30.43 \\
Gemini 2.5 Pro   & Direct Answer             & 43.73 & 32.92 & \textbf{43.81} & 45.36 & \textbf{44.54} & 36.18 & \textbf{56.52} \\
\midrule
\rowcolor{gray!15} \multicolumn{9}{c}{{\raisebox{-0.2\height}{\includegraphics[height=1.1em]{figures/logo/search.pdf}}}~~\textit{\textbf{Heuristic RAG}}} \\
\midrule
LLaMA-3.2-11B   & Image RAG               & 18.53 & 8.07  & 9.79  & 19.06 & 18.49 & 12.32 & 4.35  \\
LLaMA-3.2-11B  & Text RAG                 & 14.08 & 8.07  & 12.37 & 15.3  & 14.29 & 9.16  & 4.35  \\
LLaMA-3.2-11B  & Multimodal RAG           & 15.19 & 10.56 & 13.92 & 17.06 & 14.29 & 9.32  & 4.35  \\
Qwen2.5-VL-7B  & Image RAG                & 20.66 & 11.18 & 13.40 & 22.71 & 14.29 & 12.48 & 4.35  \\
Qwen2.5-VL-7B   & Text RAG                & 17.56 & 6.21  & 12.37 & 19.25 & 15.97 & 9.32  & 8.70  \\
Qwen2.5-VL-7B   &  Multimodal RAG         & 18.53 & 9.32  & 9.28  & 20.64 & 10.92 & 9.64  & 8.70  \\
\midrule
\textbf{\agentname{}\dag~}~(Ours) &  Multimodal RAG  
& \gaincell{38.90}{13.69}
& \gaincell{26.09}{12.43}
& \gaincell{28.87}{14.95}
& \gaincell{40.01}{13.66}
& \gaincell{30.96}{7.43}
& \gaincell{34.45}{17.55}
& \gaincell{26.09}{21.74}
\\
\midrule
\textbf{\agentname{}\ddag~}~(Ours) &  Multimodal RAG  
& \gaincell{\textbf{45.23}}{10.59}
& \gaincell{\textbf{37.27}}{13.67}
& \gaincell{37.63}{16.50}
& \gaincell{\textbf{48.76}}{11.84}
& \gaincell{33.97}{6.24}
& \gaincell{\textbf{37.82}}{14.12}
& \gaincell{21.74}{4.35}
\\
\bottomrule
\end{tabular}
}
\label{tab:dynamism-hops}
\end{table*}

%% file: tables/appendix/llm_judge.tex
\begin{table}[h]
\caption{Performance of Different Evaluators on Assessment.}
\label{tab:llm_judge}
\centering
\footnotesize
\setlength{\tabcolsep}{2pt}
\renewcommand{\arraystretch}{0.7}
\resizebox{\linewidth}{!}{
\begin{tabular}{lccccc}
\toprule
\textbf{Evaluator} & Llama-3.1-8B  & Gemma-3-27B  & Qwen2.5-32B & Hybrid LLM &  Human\\
\midrule
\textbf{Qwen2.5-VL-7B}     & 41.37 & 39.05 & 32.82  & 29.36 & 32.81 \\
\textbf{Gemini 2.5 Pro}    & 52.18 & \textbf{52.10} & 43.02  & 34.64 & 43.75 \\
\textbf{\agentname{}\ddag} & \textbf{54.75} & 51.11 & \textbf{44.10}  & \textbf{38.03} & \textbf{45.31} \\
\bottomrule
\end{tabular}
}
\vskip 0.15in
\end{table}

%% file: tables/appendix/lang_distribution.tex
\begin{table}[h]
\caption{Different Language Performance Comparison.}
\label{tab:lang_distribution}

\centering
\footnotesize
\setlength{\tabcolsep}{10pt}
\resizebox{\linewidth}{!}{
\begin{tabular}{lcccc}
\toprule
\textbf{Language}          & en (\#1369) & zh (\#880)  & fr (\#152) & jp (\#63)  \\
\midrule
\textbf{Qwen2.5-VL-7B}     & 35.06       & 32.16       & 30.26      & 36.51      \\
\textbf{Gemini 2.5 Pro}    & 44.63       & 46.82       & 37.50      & 50.79      \\
\textbf{\agentname{}\ddag} & 47.11       & 44.09       & 47.37      & 47.62      \\
\bottomrule
\end{tabular}
}
\vskip -0.15in
\end{table}

%% file: tables/appendix/datacase.tex
\begin{figure*}[h]
\begin{tcolorbox}[
    enhanced,
    width=\textwidth,
    colback=cyan!15,
    colframe=cyan!60!black,
    sharp corners,
    boxrule=0pt,
    halign=center, left=6pt,right=6pt,top=10pt,bottom=10pt
    ]
  \Large\bfseries Case Study in \ourname{}:\\
  From Campbell’s Soup Can Image to Andy Warhol’s American Nationality
\end{tcolorbox}

\begin{tcolorbox}[
    enhanced,
    colback=white,
    colframe=gray!55,
    rounded corners,
    boxrule=0.5pt,
    left=6pt,right=6pt,top=20pt,bottom=6pt]

  \centering
 
  \includegraphics[width=\linewidth]{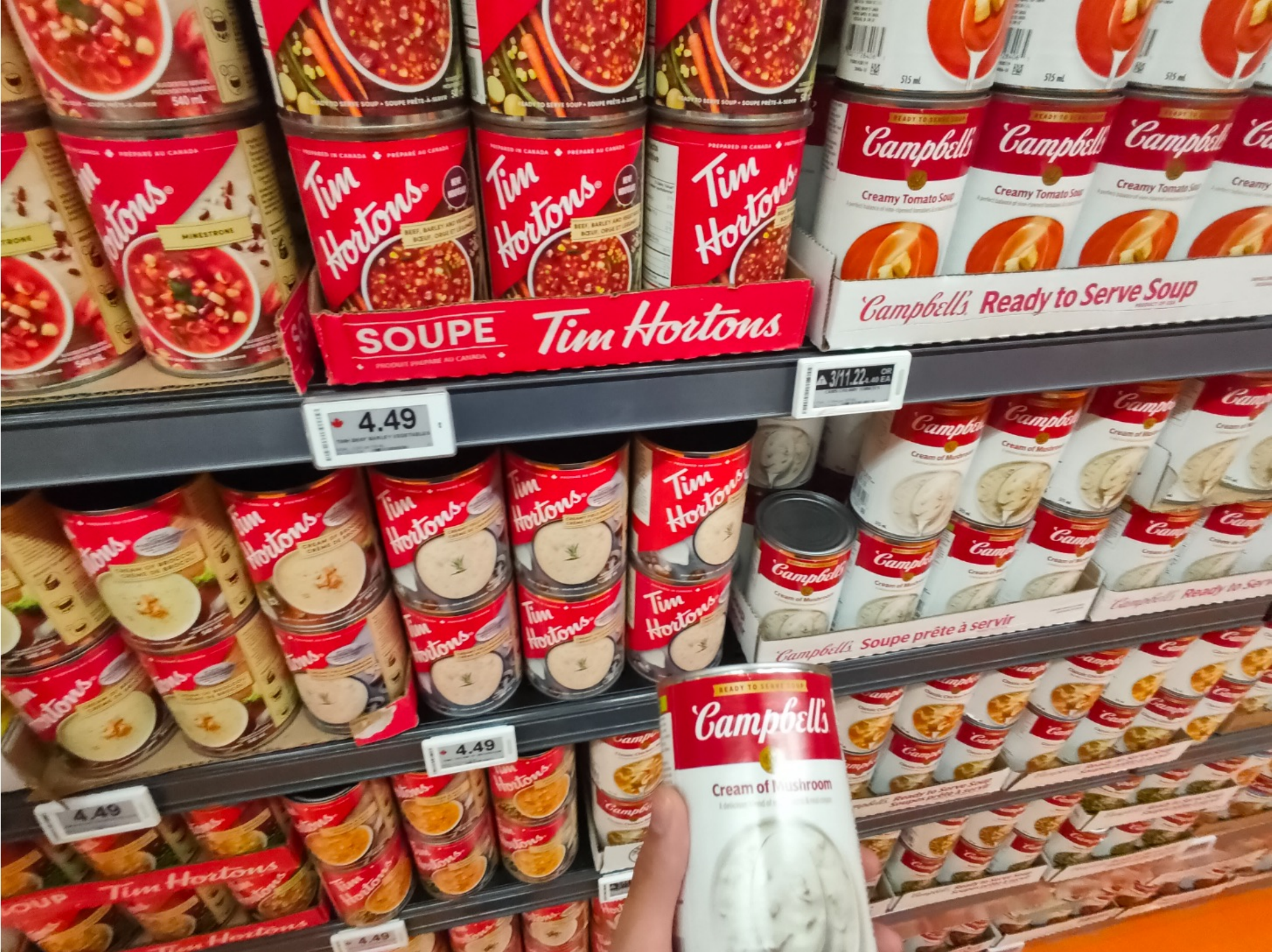}
    \vspace{20pt}
  \medskip\hrule\smallskip      
    \vspace{10pt}
  \footnotesize
  
  \centering
    \begin{tabular}{@{}l p{0.44\linewidth}@{}}
    \textbf{Question:} & Which country is the renowned artist who painted this item from? \\[4pt]
    \textbf{Answer:}   & \emph{Campbell’s} is painted by \textbf{American} pop-artist Andy Warhol. \\[6pt]

    \textbf{Glasses:}           & Xiao Mi \\
    \textbf{Image quality:}     & Normal \\
    \textbf{Domain:}            & Food \\
    \textbf{Location:}          & Canada \\
    \textbf{Category:}          & Multi-hop \\
    \textbf{Question dynamism:} & Static \\
    \textbf{Difficulty:}        & Hard \\
    \textbf{Hops number:}       & 4 \\
\end{tabular}
\vspace{10pt}
\end{tcolorbox}

\caption{Case Study in \ourname{}: ``Campbell's Soup Can''}
\label{fig:query-metadata}
\end{figure*}

\begin{figure*}[h]
\begin{tcolorbox}[title=\textbf{Hop 1},
                  colback=white,
                  colframe=black!40,
                  fonttitle=\bfseries]
  \centering
  \includegraphics[width=0.4\linewidth]{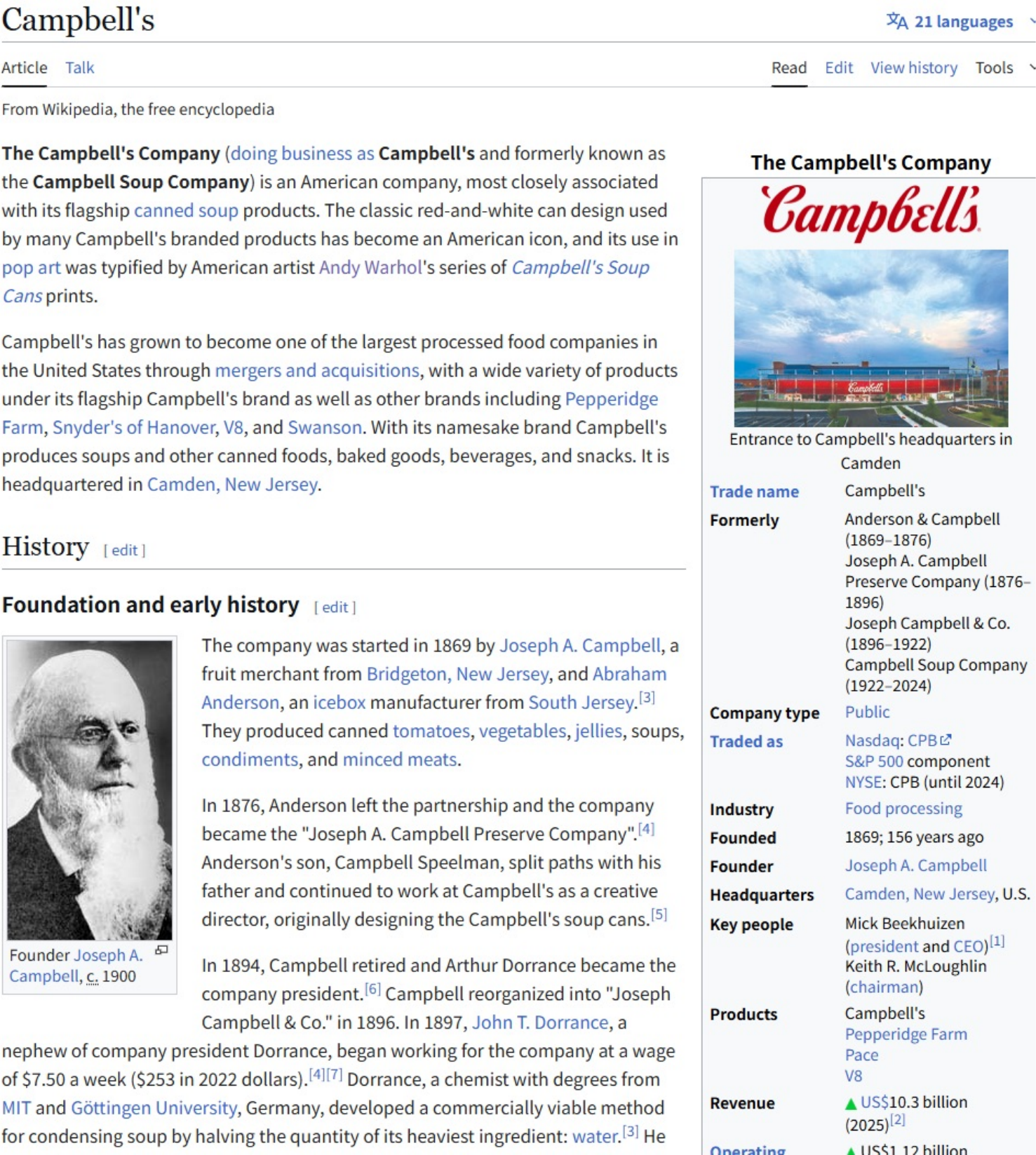} 

  \begin{flushleft}\footnotesize
    \textbf{Sub-question}: What is this product?\\[2pt]
    \textbf{Tool used}: Image Search\\[2pt]
    \textbf{Search URL}: \url{https://en.wikipedia.org/wiki/Campbell%27s}\\[2pt]
    \textbf{Search result (snippet)}: \emph{The Campbell's Company}
    (doing business as \textbf{Campbell's}) … its flagship canned
    \textbf{soup} products. 
  \end{flushleft}
\end{tcolorbox}

\begin{tcolorbox}[title=\textbf{Hop 2},
                  colback=white,
                  colframe=black!40,
                  fonttitle=\bfseries]
  \centering
  \includegraphics[width=0.4\linewidth]{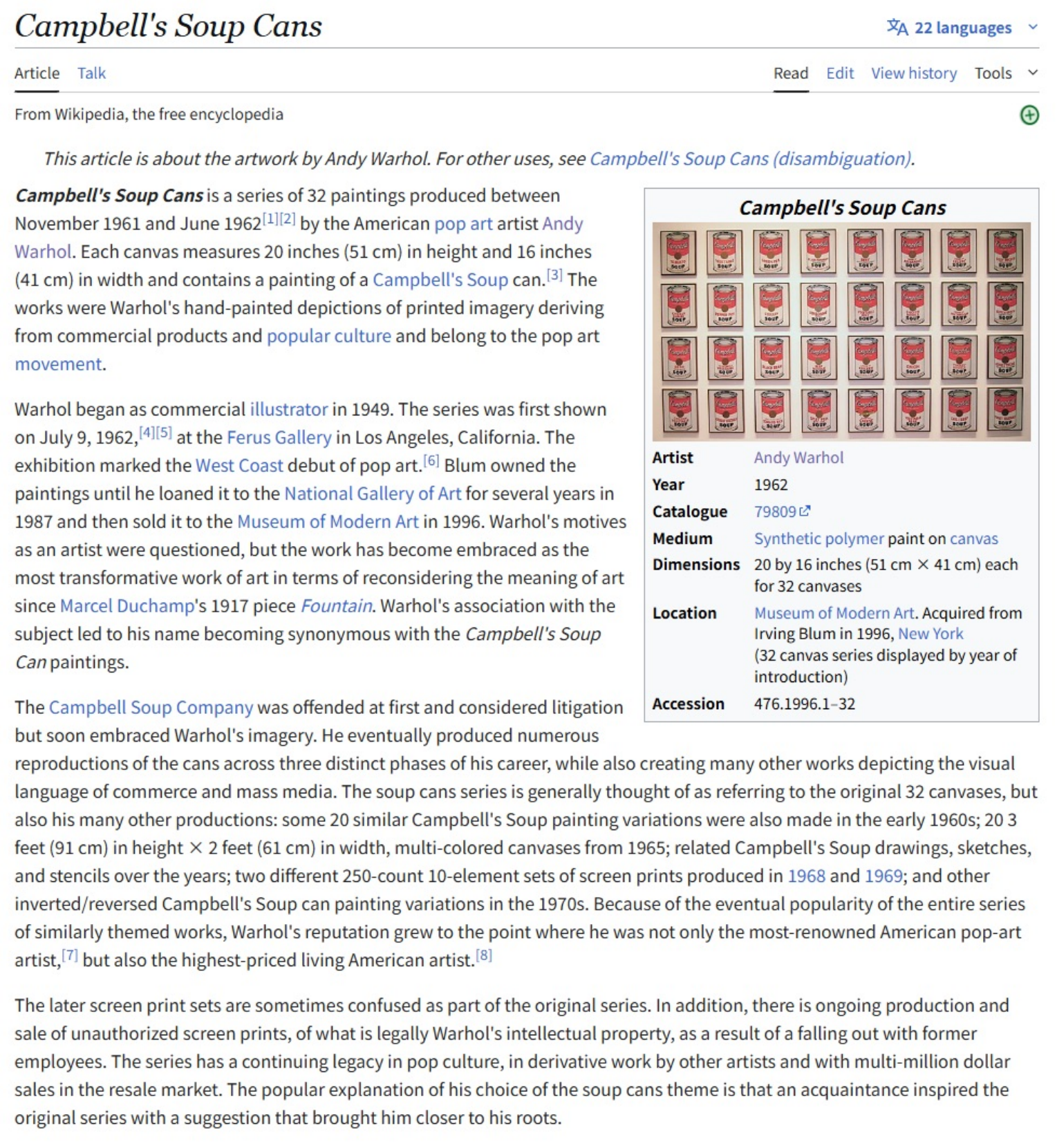}  

  \begin{flushleft}\footnotesize
    \textbf{Sub-question}: What is the famous painting depicting this item?\\[2pt]
    \textbf{Tool used}: Text Search\\[2pt]
    \textbf{Search keywords}: \emph{What is the famous painting depicting campbell?}\\[2pt]
    \textbf{Search URL}: \url{https://en.wikipedia.org/wiki/Campbell%27s_Soup_Cans}\\[2pt]
    \textbf{Search result (snippet)}: \emph{Campbell's Soup Cans} is a series of 32 paintings (1961–62) by the American pop-art artist Andy Warhol, each canvas depicting a Campbell's soup can and now considered an icon of pop art.
  \end{flushleft}
\end{tcolorbox}
\caption{Hop 1 and Hop 2 of  ``Campbell's Soup Can''}
\label{fig:hop1_hop2}
\end{figure*}

\begin{figure*}[h]
\begin{tcolorbox}[title=\textbf{Hop 3},
                  colback=white,
                  colframe=black!40,
                  fonttitle=\bfseries]
  \centering
  \includegraphics[width=0.39\linewidth]{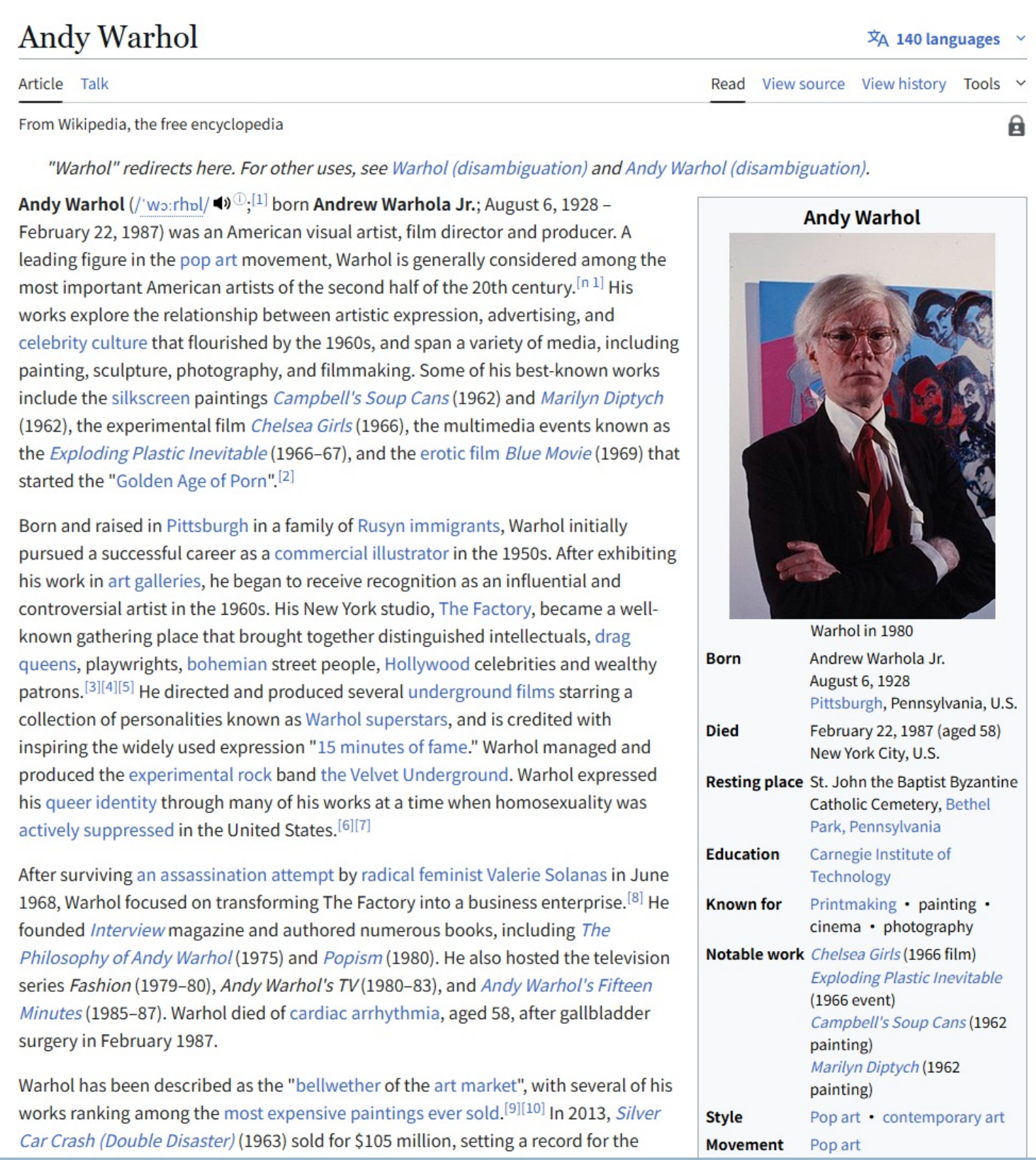}  

  \begin{flushleft}\footnotesize
    \textbf{Sub-question}: Who painted this painting?\\[2pt]
    \textbf{Tool used}: Text Search\\[2pt]
    \textbf{Search keywords}: \emph{Who paint Campbell's Soup Cans?}\\[2pt]
    \textbf{Search URL}: \url{https://en.wikipedia.org/wiki/Andy_Warhol}\\[2pt]
    \textbf{Search result (snippet)}: \textbf{Andy Warhol} was an American visual artist, film-director and producer, widely regarded as a leading figure of the pop-art movement.
  \end{flushleft}
\end{tcolorbox}

\begin{tcolorbox}[title=\textbf{Hop 4},
                  colback=white,
                  colframe=black!40,
                  fonttitle=\bfseries]
  \centering
  \includegraphics[width=0.39\linewidth]{figures/case/hop3.pdf}  

  \begin{flushleft}\footnotesize
    \textbf{Sub-question}: What nationality is he?\\[2pt]
    \textbf{Tool used}: Text Search\\[2pt]
    \textbf{Search keywords}: \emph{What nationality is Andy Warhol?}\\[2pt]
    \textbf{Search URL}: \url{https://en.wikipedia.org/wiki/Andy_Warhol}\\[2pt]
    \textbf{Search result (snippet)}: Andy Warhol was an \textbf{American} artist; he is generally considered one of the most influential U.S. figures in 20th-century art.
  \end{flushleft}
\end{tcolorbox}
\caption{Hop 3 and Hop 4 of  ``Campbell's Soup Can''}
\label{fig:hop3_hop4}
\end{figure*}